%% file: main.tex
\pdfoutput=1
\documentclass{article}

\usepackage[final]{neurips_2020}

\usepackage[utf8]{inputenc} 
\usepackage[T1]{fontenc}    
\usepackage{hyperref}       
\usepackage{url}            
\usepackage{booktabs}       
\usepackage{amsfonts}       
\usepackage{nicefrac}       
\usepackage{microtype}      
\usepackage{graphicx}
\usepackage{wrapfig}
\usepackage{subcaption}
\usepackage{xcolor}
\usepackage{soul}
\usepackage{forest}
\usepackage{tikz-qtree}
\usepackage{tikz}
\usepackage{array}
\usepackage{siunitx}
\usepackage{soul}
\usepackage{etoolbox}

\renewcommand{\bfseries}{\fontseries{b}\selectfont}
\newrobustcmd{\B}{\bfseries}

\usetikzlibrary{positioning,shapes.multipart}

\setcitestyle{numbers,square}

\usepackage{amsfonts, amsmath, amsthm, amssymb, bm, commath}
\DeclareMathOperator*{\argmin}{arg\,min}
\DeclareMathOperator*{\argmax}{arg\,max}

\newcommand{\bnb}{B$\&$B}

\title{Hybrid Models for Learning to Branch}
\author{%
Prateek Gupta\thanks{The work was done during an internship at Mila and CERC. Correspondence to: <pgupta@robots.ox.ac.uk>} \\
University of Oxford \\
The Alan Turing Institute \\
\texttt{pgupta@robots.ox.ac.uk}
\And
Maxime Gasse \\
Mila, Polytechnique Montréal \\
\texttt{maxime.gasse@polymtl.ca} 
\And
Elias B. Khalil \\
University of Toronto \\
\texttt{khalil@mie.utoronto.ca} 
\And
M. Pawan Kumar \\
University of Oxford \\
\texttt{pawan@robots.ox.ac.uk}
\And
Andrea Lodi \\
CERC, Polytechnique Montréal \\
\texttt{andrea.lodi@polymtl.ca} 
\And 
Yoshua Bengio \\
Mila, Université de Montréal \\
\texttt{yoshua.bengio@mila.quebec}
}
\begin{document}

\maketitle

\begin{abstract}

A recent Graph Neural Network (GNN) approach for learning to branch has been shown to successfully reduce the running time of branch-and-bound (\bnb{}) algorithms for Mixed Integer Linear Programming (MILP).
While the GNN relies on a GPU for inference, MILP solvers are purely CPU-based.
This severely limits its application as many practitioners may not have access to high-end GPUs. 
In this work, we ask two key questions.
First, in a more realistic setting where only a CPU is available, is the GNN model still competitive?
Second, can we devise an alternate computationally inexpensive model that retains the predictive power of the GNN architecture?
We answer the first question in the negative, and address the second question by proposing a new hybrid architecture for efficient branching on CPU machines. 
The proposed architecture combines the expressive power of GNNs with computationally inexpensive multi-layer perceptrons (MLP) for branching.
We evaluate our methods on four classes of MILP problems, and show that they lead to up to 26\% reduction in solver running time compared to state-of-the-art methods without a GPU, while extrapolating to harder problems than it was trained on. The code for this project is publicly available at \url{https://github.com/pg2455/Hybrid-learn2branch}.
\end{abstract}

\section{Introduction}

Mixed-Integer Linear Programs (MILPs) arise naturally in many decision-making problems such as auction design~\citep{abrache2007combinatorial}, warehouse planning~\citep{elson1972site}, capital budgeting~\citep{capbudgeting} or scheduling~\citep{floudas2005mixed}. 
Apart from a linear objective function and linear constraints, some decision variables of a MILP are required to take integral values, which makes the problem NP-hard~\citep{book/PapadimitriouS82}.

Modern mathematical solvers typically employ the \bnb{} algorithm~\citep{journal/econometrica/LandD60} to solve general MILPs to global optimality. 
While the worst-case time complexity of \bnb{} is exponential in the size of the problem ~\citep{books/Wolsey98}, it has proven efficient in practice, leading to wide adoption in various industries.
At a high level, \bnb{} adopts a divide-and-conquer approach that consists in recursively partitioning the original problem into a tree of smaller sub-problems, and solving linear relaxations of the sub-problems until an integral solution is found and proven optimal.

Despite its apparent simplicity, there are many practical aspects that must be considered for \bnb{} to perform well~\citep{achterberg_thesis}; such decisions will affect the search tree, and ultimately the overall running time. 
These include several decision problems~\citep{journals/top/LodiZ2017} that arise during the execution of the algorithm, such as \textit{node selection}: which sub-problem do we analyze next?; and \textit{variable selection} (a.k.a. branching): which decision variable must be used (branched on) to partition the current sub-problem?
While such decisions are typically made using hard-coded expert heuristics which are implemented in modern solvers, more and more attention is given to statistical learning approaches for replacing and improving upon those heuristics \citep{conf/nips/HeDE14,Alvarez2014ASM, confs/aaai/KhalilBND16,gasse2019exact,zarpellon2020parameterizing}. 
An extensive review of different approaches at the intersection of statistical learning and combinatorial optimization is given in \citet{arxiv/BengioLP18}.

\input{inputs/times_arch}
Recently, \citet{gasse2019exact} proposed to tackle the variable selection problem in \bnb{} using a Graph Neural Network (GNN) model.
The GNN exploits the bipartite graph formulation of MILPs together with a shared parametric representation, thus allowing it to model problems of arbitrary size.
Using imitation learning, the model is trained to approximate a very good but computationally expensive ``expert" heuristic named \textit{strong branching} \citep{techreport/ApplegateBCC95}. 
The resulting branching strategy is shown to improve upon previously proposed approaches for branching on several MILP problem benchmarks, and is competitive with state-of-the-art \bnb{} solvers.
We note that one limitation of this approach, with respect to general \bnb{} heuristics, is that the resulting strategy is only tailored to the class of MILP problems it is trained on. 
This is very reasonable in our view, as practitioners usually only care about solving very specific problem types at any time.

While the GNN model seems particularly suited for learning branching strategies, one drawback is a high computational cost for inference, i.e., choosing the branching variable at each node of the \bnb{} tree. 
In~\citet{gasse2019exact}, the authors use a high-end GPU card to speed-up the GNN inference time, which is a common practice in deep learning but is somewhat unrealistic for MILP practitioners. 
Indeed, commercial MILP solvers rely solely on CPUs for computation, and the GNN model from \citet{gasse2019exact} is not competitive on CPU-only machines, as illustrated in Figure~\ref{fig:times_arch}. 
There is indeed a trade-off between the quality of the branching decisions made and the time spent obtaining those decisions.
This trade-off is well-known in MILP community~\citep{achterberg_thesis}, and has given rise to carefully balanced strategies designed by MILP experts, such as \textit{hybrid branching}~\citep{confs/cpaior/AchterbergB09}, which derive from the computationally expensive \textit{strong branching} heuristic~\citep{techreport/ApplegateBCC95}.

In this paper, we study the time-accuracy trade-off in learning to branch with the aim of devising a model that is both computationally inexpensive and accurate for branching.
To this end, we propose a hybrid architecture that uses a GNN model only at the root node of the \bnb{} tree and a weak but fast predictor, such as a simple Multi-Layer Perceptron (MLP), at the remaining nodes.
In doing so, the weak model is enhanced by high-level structural information extracted at the root node by the GNN model.
In addition to this new hybrid architecture, we experiment and evaluate the impact of several variants to our training protocol for learning to branch, including: (i) end-to-end training~\citep{glasmachers2017limits, bojarski2016end}, (ii) knowledge distillation~\citep{hinton2015distilling}, (iii) auxiliary tasks~\citep{liebel2018auxiliary}, and (iv) depth-dependent weighting of the training loss for learning to branch, an idea originally proposed by  \citet{conf/nips/HeDE14} in the context of node selection.

We evaluate our approach on large-scale MILP instances from four problem families: Capacitated Facility Location, Combinatorial Auctions, Set Covering, and Independent Set.
We demonstrate empirically that our combination of hybrid architecture and training protocol results in state-of-the-art performance in the realistic setting of a CPU-restricted machine. 
While we observe a slight decrease in the predictive performance of our model with respect to the original GNN from~\citep{gasse2019exact}, its reduced computational cost still allows for a reduction of up to 26\% in overall solving time on all the evaluated benchmarks compared to the default branching strategy of the modern open-source solver SCIP~\citep{GleixnerEtal2018OO}. 
Also, our hybrid model preserves the ability to extrapolate to harder problems than trained on, as the original GNN model.

\section{Related Work}

Finding a branching strategy that results in the smallest \bnb{} tree for a MILP is at least as hard--and possibly much harder--than solving the MILP with any strategy. 
Still, small trees can be obtained by using a computationally expensive heuristic named \textit{strong branching} (SB)~\citep{techreport/ApplegateBCC95, lindSB}.
The majority of the research efforts in variable selection are thus aimed at matching the performance of SB through faster approximations, via cleverly handcrafted heuristics such as \textit{reliability pseudocost branching}~\citep{ACHTERBERG200542}, and recently via machine learning\footnote{Interestingly, early works on ML methods for branching can be traced back to 2000 \citep[Acknowledgements]{achterberg_thesis}.}~\citep{Alvarez2014ASM,confs/aaai/KhalilBND16, gasse2019exact}. We refer to~\cite{journals/top/LodiZ2017} for an extensive survey of the topic.

\citet{Alvarez2014ASM} and \citet{confs/aaai/KhalilBND16} showed that a fast discriminative classifier such as extremely randomized trees~\citep{geurts2006extremely} or support vector machines~\citep{svmHearst} on hand-designed features can be used to mimic SB decisions.
Subsequently, \citet{gasse2019exact} and \citet{zarpellon2020parameterizing} showed the importance of representation learning for branching.
Our approach, in some sense, combines the superior representation framework of \citet{gasse2019exact} with the computationally cheaper framework of \citet{confs/aaai/KhalilBND16}. 
Such hybrid architectures have been successfully used in ML problems such as visual reasoning~\citep{perez2018film}, style-transfer~\citep{dumoulin2016learned}, natural language processing~\citep{srivastava2015highway, dhingra2016gated}, and speech recognition~\citep{kim2017dynamic}.

\section{Preliminaries}

Throughout this paper, we use boldface for vectors and matrices. A MILP is a mathematical optimization problem that combines a linear objective function, a set of linear constraints, and a mix of continuous and integral decision variables. It can be written as:
\begin{equation*}
    \argmin_{\mathbf{x}} \mathbf{c}^{\intercal}\mathbf{x}, \quad \text{s.t.} \quad  \mathbf{A}\mathbf{x} \leq \mathbf{b}, \quad  \mathbf{x} \in \mathbb{Z}^{p} \times \mathbb{R}^{n-p},
\end{equation*}
where $\mathbf{c} \in \mathbb{R}^n$ denotes the cost vector, $\mathbf{A} \in \mathbb{R}^{m\times n}$ the matrix of constraint coefficients, $\mathbf{b} \in \mathbb{R}^m$ is the vector of constant terms of the constraints, and there are $p$ integer variables, $1\leq p \leq n$ .

The \bnb{} algorithm can be described as follows.
One first solves the linear program (LP) relaxation of the MILP, obtained by disregarding the integrality constraints on the decision variables.
If the LP solution $\mathbf{x}^\star$ satisfies the MILP integrality constraints, or is worse than a known integral solution, then there is no need to proceed further. 
If not, then one divides the MILP into two sub-MILPs. This is typically done by picking an integral decision variable that has a fractional value, $i \in \mathcal{C}=\{i \mid x^\star_i \not\in \mathbb{Z}, i \leq p\}$, and create two sub-MILPs with additional constraints $x_i \leq \lfloor x^\star_i \rfloor$ and $x_i \geq \lceil x^\star_i \rceil$, respectively. 
The decision variable $i$ that is used to partition the feasible region is called the \textit{branching variable}, while $\mathcal{C}$ denotes the \textit{branching candidates}. 
The second step is to select one of the leaves of the tree, and repeat the above steps until all leaves have been processed\footnote{For a more involved description of \bnb{}, the reader is referred to \citet{achterberg_thesis}.}.

In this work, we refer to the first node processed by \bnb{} as the \textit{root node}, which contains the original MILP, and all subsequent nodes containing a local MILP as \textit{tree nodes}, whenever the distinction is required. Otherwise we refer to them simply as nodes.

\section{Methodology}
\label{sec:arch&train}

As mentioned earlier, computationally heavy GNNs can be prohibitively slow when used for branching on CPU-only machines.
In this section we describe our hybrid alternative, which combines the superior inductive bias of a GNN at the root node with a computationally inexpensive model at the tree nodes. We also discuss various enhancements to the training protocol, in order to enhance the performance of the learned models.

\subsection{Hybrid architecture}

A variable selection strategy in \bnb{} can be seen as a scoring function $f$ that outputs a score $s_i \in \mathbb{R}$ for every branching candidate. As such, $f$ can be modeled as a parametric function, learned by ML. 
Branching then simply involves selecting the highest-scoring candidate according to $f$:
\begin{equation*}
    \qquad i^\star_f = \argmax_{i \in \mathcal{C}} \mathbf{s}_i
\end{equation*}

We consider two forms of node representations for machine learning models: (i) a graph representation $\mathbf{G} \in \mathcal{G}$, such as the variable-constraint bipartite graph of \citet{gasse2019exact}, where $\mathbf{G} = (\mathbf{V}, \mathbf{E}, \mathbf{C})$, with $\mathbf{V} \in \mathbb{R}^{n \times d_1}$ variable features, $\mathbf{E} \in \mathbb{R}^{n \times m \times d_2}$ edge features, and $\mathbf{C} \in \mathbb{R}^{m \times d_3}$ constraint features; and (ii) branching candidate features
$\mathbf{X} \in \mathbb{R}^{|\mathcal{C}| \times d_4}$, such as those from \citet{confs/aaai/KhalilBND16}, which is cheaper to extract than the first representation. 
For convenience, we denote by $\mathcal{X}$ the generic space of the branching candidate features, and by $\mathbf{G}^0$ the graph representation of the root node.
The various features $d_i$ used in this work are detailed in the supplementary materials.

In \bnb{}, structural information in the tree nodes, $\mathbf{G}$, shares a lot of similarity with that of the root node, $\mathbf{G}^0$. Extracting, but also processing that information at every node is an expensive task, which we will try to circumvent. The main idea of our hybrid approach is then to succinctly extract the relevant structural information only once, at the root node, with a parametric model $\text{GNN}(\mathbf{G}^0; \pmb{\theta})$. We then combine in the tree nodes this preprocessed structural information with the cheap candidate features $\mathbf{X}$, using a hybrid model $f := \text{MLP}_\text{HYBRID}(\text{GNN}(\mathbf{G}^0; \pmb{\theta}), \mathbf{X}; \pmb{\phi})$. By doing so, we hope that the resulting model will approach the performance of an expensive but powerful $f := \text{GNN}(\mathbf{G})$, at almost the same cost as an inexpensive but less powerful $f := \text{MLP}(\mathbf{X})$. Figure~\ref{fig:arch-data-type} illustrates the differences between those approaches, in terms of data extraction.

\input{inputs/arch-data-types}

For an exhaustive coverage of the computational spectrum of hybrid models, we consider four ways to enrich the feature space of an MLP via a GNN's output, sumarrized in Table~\ref{tabl:arch-types}.
In CONCAT, we concatenate the candidate's \textit{root representations} $\pmb{\Psi}$ with the features $\mathbf{X}$ at a node.
In FiLM~\citep{perez2018film}, we generate film parameters $\pmb{\gamma}$, $\pmb{\beta}$ from the GNN, for each candidate, which are further used to modulate the hidden layers of the MLP. In details, if $\pmb{h}$ is the intermediate representation of the MLP, it gets linearly modulated as $\pmb{h} \leftarrow \pmb{\beta} \cdot  \pmb{h} + \pmb{\gamma}$.
While both the above architectures have similar computational complexity, it has been shown that FiLM subsumes the CONCAT architecture~\cite{dumoulin2018feature}.
On the other the end of the spectrum lie the most inexpensive hybrid architectures, HyperSVM and HyperSVM-FiLM.
HyperSVM is inspired by HyperNetworks~\citep{ha2016hypernetworks}, and simply consists in a multi-class Support Vector Machine (SVM), whose parameters are predicted by the root GNN. 
We chose a simple linear disciminator, for a minimal computational cost. 
Finally, in HyperSVM-FiLM we increase the expressivity of HyperSVM with the help of modulations, similar to that of FiLM. 
\input{inputs/arch-types}

\subsection{Training Protocol}
We use \textit{strong branching} decisions as ground-truth labels for imitation learning, and collect observations of the form $(\mathbf{G}^0, \mathbf{G}, \mathbf{X}, i^\star_{SB})$.
Thus, the data used for training the model is $\mathcal{D} = \{(\mathbf{G}^0_k, \mathbf{G}_k, \mathbf{X}_k, i^\star_{SB,k}), k=1,2,...,N\}$.
We treat the problem of identifying $i^\star_{SB,k}$ as a classification problem, such that $i^* = \argmax_{i \in \mathcal{C}} f(\mathbf{G}^0, \mathbf{X})$ is the target outcome.
Considering $\mathcal{D}$ as our ground-truth, our objective (\ref{eq:loss-fn}) is to minimize the cross-entropy loss $l$ between $f(\mathbf{G}^0, \mathbf{X}) \in \mathcal{R}^{|\mathcal{C}|}$ and
a one-hot vector with one at the target:
\begin{equation}
   \mathcal{L}(\mathcal{D}; \pmb{\theta}, \pmb{\phi}) = \frac{1}{N}\sum_{k=1}^N l(f(\mathbf{G}^0_k, \mathbf{X}_k ; \pmb{\theta}, \pmb{\phi}), i^\star_{SB,k}).
    \label{eq:loss-fn}
\end{equation}

Performance on unseen instances, or generalization, is of utmost importance when the trained models are used on bigger instances.
The choices in training the aforementioned architectures influence this ability.
In this section, we discuss four such important choices that lead to better generalization.

\subsubsection{End-to-end Training (e2e)}
A $\text{GNN}_{\pmb{\hat{\theta}}}$, with pre-trained parameters $\pmb{\hat{\theta}}$, is obtained using the procedure described in \citet{gasse2019exact}. 
We use this pre-trained GNN to extract variable representations at the root node and use it as an input to the MLPs at a tree node (pre).
This results in a considerable performance boost over plain "salt-of-the-earth" MLP used at all tree nodes. 
However, going a step further, an end-to-end (e2e) approach involves training the GNN and the MLP together by backpropagating the gradients from a tree node to the root node.
In doing so, e2e training aligns the variable representations at the root node with the prediction task at a tree node.
At the same time, it is not obvious that it should result in a stable learning behavior because the parameters for GNN need to adapt to various tree nodes.
Our experiments explore both pre-training (pre) and end-to-end training (e2e), namely:
\begin{equation}\label{eq:e2e}
    \text{(pre)} \qquad \pmb{\phi}^* = \argmin_{\pmb{\phi}}\mathcal{L}(\mathcal{D}; \pmb{\hat{\theta}}, \pmb{\phi} )
    \text{,} \qquad
    \text{(e2e)} \qquad \pmb{\phi}^*, \pmb{\theta}^* = \argmin_{\pmb{\theta}, \pmb{\phi}}\mathcal{L(\mathcal{D}; \pmb{\theta}, \pmb{\phi})}.
\end{equation}

\subsubsection{Knowledge Distillation (KD)}
Using the outputs of a pre-trained expert model as a soft-target for training a smaller model has been successfully used in model compression~\citep{hinton2015distilling}.
In this way, one aims to learn an inexpensive model that has the same generalization power as an expert.
Thus, for better generalization, instead of training our hybrid architectures with cross-entropy~\citep{Goodfellow-et-al-2016} on ground-truth hard-labels, we study the effect of training with KL Divergence~\citep{kullback1951information} between the outputs of a pre-trained GNN and a hybrid model, namely:
\begin{equation}\label{eq:kd}
   \text{(KD)} \qquad \mathcal{L}_{KD}(\mathcal{D}; \pmb{\theta}, \pmb{\phi}) = \frac{1}{N}\sum_{k=1}^N \text{KL}(f(\mathbf{G}^0_k, \mathbf{X}_k ; \pmb{\theta}, \pmb{\phi}), \text{GNN}_{\pmb{\hat{\theta}}}(\mathbf{G_k})). 
\end{equation}

\subsubsection{Auxiliary Tasks (AT)}
An inductive bias, such as GNN, encodes a prior on the way to process raw input.
Auxiliary tasks, on the other hand, inject priors in the model through additional learning objectives, which are not directly linked to the main task.
These tasks are neither related to the final output nor do they require additional training data.
One such auxiliary task is to maximize the diversity in variable representations. 
The intuition is that very similar representations lead to very close MLP score predictions, which is not useful for branching.

We minimize a pairwise loss function that ensures maximum separation between the variable representations projected on a unit hypersphere.
We consider two types of objectives for this: (i) Euclidean Distance (ED), and (ii) Minimum Hyperspherical Energy (MHE)~\citep{liu2018learning}, inspired from the well-known Thomson problem~\cite{thomson} in Physics.
While ED separates the representations in the Euclidean space on the hypersphere, MHE ensures uniform distribution over the hypersphere.
Denoting $\hat{\pmb{\psi}_i}$ as the variable representation for the variable $i$ projected on a unit hypersphere and $e_{ij} = ||{\hat{\pmb{\psi}_i} - \hat{\pmb{\psi}_j}}||_2$ as the Euclidean distance between the representations for the variables $i$ and $j$, our new objective function is given as $\mathcal{L}_{AT}(\mathcal{D};\pmb{\theta}, \pmb{\phi}) = \mathcal{L}(\mathcal{D};\pmb{\theta}, \pmb{\phi}) + g(\mathbf{\Psi};\pmb{\theta})$, where
\begin{equation}\label{eq:at}
    \text{(ED)} \qquad  g(\mathbf{\Psi};\pmb{\theta}) = \frac{1}{N^2}\sum_{i,j=1}^{N} e_{ij}^2
    \text{,} \qquad
    \text{(MHE)} \qquad g(\mathbf{\Psi};\pmb{\theta}) = \frac{1}{N^2}\sum_{i,j=1}^{N}\frac{1}{e_{ij}}.
\end{equation}

\subsubsection{Loss Weighting Scheme}
\label{sec:weighting}
The problem of distribution shift is unavoidable in a sequential process like B\&B. 
A suboptimal branching decision at a node closer to the root node can have worse impact on the size of the \bnb{} tree as compared to when such a decision is made farther from it.
In such situations, one can use depth (possibly normalized) as a feature, but the generalization on bigger instances is a bit unpredictable as the distribution of this feature might be very different from that observed in the training set.
Thus, we experimented with different depth-dependent formulations for weighting the loss at any node.
Denoting $z_i$ as the depth of a tree node $i$ relative to the depth of the tree, we weight the loss at different tree nodes by $w(z_i)$, making our objective function as 
\begin{equation}\label{eq:weights}
   \mathcal{L}(\mathcal{D}; \pmb{\theta}, \pmb{\phi}) = \frac{1}{N}\sum_{k=1}^N w(z_k) \cdot  l(f(\mathbf{G}^0_k, \mathbf{X}_k ; \pmb{\theta}, \pmb{\phi}), i^\star_{SB,k}).
\end{equation}

Specifically, we considered 5 different weighting functions such that all of them have the same end points, i.e., $w(0) = 1.0$ at the root node and $w(1) = e^{-0.5}$ at the deepest node.
Different functions were chosen depending on their intermediate behaviour in between these two points. 
We experimented with exponential, linear, quadratic and sigmoidal decay behavior of these functions. 
Table~\ref{tab:weights} lists various functions and their mathematical forms considered in our experiments. 

\section{Experiments}
\label{sec:experiments}
We follow the experimental setup of \citet{gasse2019exact}, and evaluate each branching strategy across four different problem classes, namely Capacitated Facility Location, Minimum Set Covering, Combinatorial Auctions, and Maximum Independent Set. 
Randomly generated instances are solved offline using SCIP~\citep{GleixnerEtal2018OO} to collect training samples of the form $(\mathbf{G}^0, \mathbf{G}, \mathbf{X}, i^\star_{SB})$. 
We leave the description of the data collection and training details of each model to the supplementary materials.

\paragraph{Evaluation.} As in \citet{gasse2019exact}, our evaluation instances are labeled as small, medium, and big based on the size of underlying MILP. Small instances have the same size as those used to generate the training datasets, and thus match the training distribution, while instances of increasing size allows us to measure the generalization ability of the trained models. Each scenario uses 20 instances, solved using 3 different seeds to account for solver variability.
We report standard metrics used in the MILP community for benchmarking \bnb{} solvers: (i) Time: 1-shifted geometric mean\footnote{for complete definition refer to Appendix A.3 in ~\citet{achterberg_thesis}} of running times in seconds, including the running times for unsolved instances, (ii) Nodes: hardware-independent 1-shifted geometric mean of \bnb{} node count of the instances solved by all branching strategies
, and (iii) Wins: number of times each branching strategy resulted in the fastest solving time, over total number of solved instances.
All branching strategies are evaluated using the open-source solver SCIP~\citep{GleixnerEtal2018OO} with a time limit of 45 minutes, and cutting planes are allowed only at the root node.

\input{inputs/model_results}
\paragraph{Baselines.} We compare our hybrid model to several non-ML baselines, including SCIP's default branching heuristic \textit{Reliability Pseudocost Branching} (RPB), the ``gold standard" heuristic \textit{Full Strong Branching} (FSB), and a very fast but ineffective heuristic \textit{Pseudocost Branching} (PB).
We also include the GNN model from \citet{gasse2019exact} run on CPU (GNN), and also several fast but less expressive models such as SVMRank from \citet{confs/aaai/KhalilBND16}, LambdaMART from  \citet{burges2010ranknet}, and ExtraTree Classifier from \citet{geurts2006extremely} as benchmarks. 
For conciseness, we chose to report those last three competitor models using an optimistic aggregation scheme, by systematically choosing only the best performing method among the three (COMP).
For completeness, we also report the performance of a GNN model run on a high-end GPU, although we do not consider that method as a baseline and therefore do not include it in the Wins indicator.
We acknowledge that our comparison to general branching strategy like RPB is not completely fair, however, developing specialized versions of such strategies is a challenge in itself full of modeling choices that we foresee as future work.

\paragraph{Model selection.} To investigate the effectiveness of different architectures, we empirically compare the performance of their end-to-end variants.
Figure~\ref{fig:model_results} compares the Top-1 test accuracy of models across the four problem sets.
The performance of GNNs (blue), being the most expressive model, serves as an upper bound to the performance of hybrid models.
All of the considered hybrid models outperform MLPs (red) across all problem sets.
Additionally, we observe that FiLM (green) and CONCAT (purple) perform significantly better than other architectures.
However, there is no clear winner among them.
We also note that the cheapest hybrid models, HyperSVM and HyperSVM-FiLM, though better than MLPs, still do not perform as well as FiLM or CONCAT models.

\paragraph{Training protocols.} In Table~\ref{tabl:protocol-results}, we show the effect of different protocols discussed in section~\ref{sec:arch&train} on Top-1 accuracy of the FiLM models.
We observe that the presence of these protocols improves the model accuracy by 0.5-0.9\%, which translates to a minor yet practically useful improvement in \bnb{} performance of the solver.
Except for Combinatorial Auctions, FiLM's performance is improved by knowledge distillation, which suggests that the soft targets of pre-trained GNN yield a better generalization performance.
Lastly, we launch a hyperparameter search for auxiliary objective: ED and MHE, on top of the best performing model (Top-1 accuracy), among e2e and e2e \& KD models.
Auxiliary tasks further help the accuracy of the hybrid models, but for some problem classes it is ED that works well while for others it is MHE.
We provide the model performances on test dataset in the supplement for further reference.

\input{inputs/combined-protocol-loss-weighting}
\paragraph{Effect of loss weighting.} We empirically investigate the effect of the different loss weighting schemes discussed in Section~\ref{sec:weighting}. We train a simple MLP model on our small Combinatorial Auctions instances, and measure the resulting \bnb{} tree size on big instances. We report aggregated results in Table~\ref{tab:weights}, and provide instance-level results in the supplement.
We observe that the most commonly used exponential and linear schemes actually seem to degrade the performance of the learned strategy, as we believe those may be too aggressive at disregarding nodes early on in the tree. On the other hand, both the quadratic and sigmoidal schemes result in an improvement, thus validating the idea that depth-dependent weighting can be beneficial for learning to branch. We therefore opt for a sigmoidal loss weighting scheme in our training protocol.

\paragraph{Complete benchmark.} Finally, to evaluate the runtime performance of our hybrid approach, we replace  SCIP's default branching strategy with our best performing model, FiLM.
We observe in Table~\ref{tab:evaluation} that FiLM performs substantially better than all other CPU-based branching strategies.
The computationally expensive FSB, our ``gold standard'', becomes impractical as the size of instances grows, whereas RPB remains competitive. 
While GNN retains its small number of nodes, it loses in running time performance on CPU.
Note that we found that FiLM models for Maximum Independent Set did initially overfit on small instances, such that the performance on larger instances degraded substantially.
To overcome this issue we used weight decay~\cite{ng2004feature} regularization, with a validation set of 2000 observations generated using random medium instances (not used for evaluation).
We report the performance of the regularized models in the supplement, and use the best performing model to report evaluation performance on medium and big instances.
We also show in the supplement that the cheapest computation model of HyperSVM/HyperSVM-FiLM do not provide any runtime advantages over FiLM.
We achieve up to 26\% reduction on medium instances and up to 8\% reduction on big instances in overall solver running time compared to the next-best branching strategy, including both learned and classical strategies. 
\input{inputs/evaluation-compressed-format}

Finally, we note that the majority of ``big" problems in Set Covering and Maximum Independent Set are not solved by any of the branching strategies. 
\input{inputs/optimality-gap}
Therefore, we provide a comparison of optimality gap of unsolved instances at time out in Table~\ref{table:opt-gap}.
It is evident that FiLM models are able to close a larger optimality gap than the other branching strategies.

In the absence of significant difference in KD and KD \& AT model's Top-1 accuracy (see Table~\ref{tabl:protocol-results}), a natural question then to ask is: what can be a practically useful choice?
To answer this question, we tested the effect of each training protocol on the final \bnb{} performance. 
Although the detailed discussion is in the supplement we conclude that in the absence of significant difference in the test accuracy of models between KD and AT \& KD, a preference should be made for KD models to attain better generalization.

\paragraph{Limitations.} We would like to point out some limitations of our work. 
First, given the NP-Hard nature of MILP solving, it is fairly time consuming to evaluate performance of the trained models on the instances bigger than considered for this work.
One can consider the primal-dual bound gap after a time limit as an evaluation metric for the bigger instances, but this is misaligned with the solving time objective. 
Second, we have used Top-1 accuracy on the test set as a proxy for the number of nodes, but there is an inherent distribution shift because of the sequential nature of \bnb{} that leads to out-of-distribution observations.
Third, generalization to larger instances is a central problem in the design of branching strategies.
Several techniques discussed in this work form only a part of the solution to this problem.
On the Maximum Independent Set problem, we originally noticed a poor generalization capability, which we addressed by cross-validation using a small validation set. 
In future work, we plan to perform an extensive study of the effect of architecture and training protocols on generalization performance.
Another experiment worth conducting would be to train on larger instances than ``small" problems used in the work, in order to get a better view of how and to which point the different models are able to generalize.
However, not only is this a time-consuming process, but there is also an upper limit on the size of the problems on which it is reasonable to conduct experiments, simply due to hardware constraints. 
Finally, although we showed the efficacy of our models on a broad class of MILPs, there may be other problem classes for which our models might not result in a substantial runtime improvements. 

\section{Conclusion}
As more operations research and integer programming tools start to include ML and deep learning modules, it is necessary to be mindful of the practical bottlenecks faced in that domain. To this end, we combine the expressive power of GNNs with the computational advantages of MLPs to yield novel hybrid models for learning to branch, a central component in MILP solving. 
We integrate various training protocols that augment the basic MLPs and help bridge the accuracy gap with more expensive models. 
This competitive accuracy translates into savings in time and nodes when used in MILP solving, as compared to both default, expert-designed branching strategies and expensive GNN models, thus obtaining the ``best of both worlds" in terms of the time-accuracy trade-off.
More broadly, our philosophy revolves around understanding the intricacies and practical constraints of MILP solving and carefully adapting deep learning techniques therein. We believe that this integrative approach is crucial to the adoption of statistical learning in exact optimization solvers. 

\section*{Broader Impact}

This paper establishes a bridge between the work done in ML in the last years on learning to branch and the traditional MILP solvers used to routinely solve thousands of optimization applications in energy, telecommunications, logistics, biology, just to mention a few. 

The MILP solvers are executed on CPU-only machines and the technical challenge of using GPU-based algorithmic techniques to hybridize them had been neglected thus far. Admittedly, such a challenge was not urgent when the learning to branch literature was in its early stages. That situation has changed drastically with the GNN implementation in \cite{gasse2019exact}, the first approach to show significant benefit with respect to the default version of a state-of-the-art MILP solver like SCIP. For this reason, the current paper comes at the due time for the literature in the field and addresses the challenge, for the first time, in a sophisticated, yet relatively simple way. 

Thus, our work provides the first viable way for commercial and noncommercial MILP solver developers to implement and integrate a ML-based ``learning to branch" framework and for hundreds of thousands of users and practitioners to use it. In an even broader sense, the fact that we were able to approximate the performance of GPU-based models with a sophisticated integration of CPU-based techniques is  consistent with, for example, \citet{hinton2015distilling}, and widens the space of problems to which ML techniques can be successfully applied.

\begin{ack}
The authors are grateful to CIFAR and IVADO for funding and Compute Canada for computing resources.

We would further like to acknowledge the important role played by our colleagues at Mila and CERC through building a fun learning environment. 
We would also like to thank Felipe Serano and Benjamin M\"uller for their technical help with SCIP and insightful discussions on branching in MILPs. 
PG wants to thank Giulia Zarpellon, Didier Chételat, Antoine Prouvost, Karsten Roth, David Yu-Tung Hui, Tristan Deleu, Maksym Korablyov, and Alex Lamb for enlightening discussions on deep learning and integer programming. 
\end{ack}

\small
\bibliography{references}

\end{document}


\maketitle

\section{Inefficiency in using GNNs for solving MILPs in parallel}
In this section, we argue that the GNN architecture looses its advantages in the face of solving multiple MILPs at the same time.
In the applications like multi-objective optimization~\citep{kirlik2014new}, where multiple MILPs are solved in parallel, a GNN for each MILP needs to be initialized on the GPU because of the \textit{sequentially asynchronous} nature of solving MILPs.
Not only is there a limit to the number of such GNNs that can fit on a single GPU because of memory constraints, but also several GNNs on a single GPU results in an inefficient GPU utilization. 

One can, for instance, try to time multiple MILPs such that there is a need for a single forward evaluation on a GPU, but, in our knowledge, it has not been done and it results in frequent interruptions in the solving procedure.
An alternative, much simpler, method is to \textit{pack} multiple GNNs on a single GPU such that each GNN is dedicated to solving one MILP. 
For example, we were able to put 25 GNNs on Tesla V100 32 GB GPU.
Figure~\ref{fig:batch_pack} shows the inefficient utilization of GPUs when multiple GNNs are packed on a single GPU. 

\input{inputs/batch_pack}

\section{Input Features}
We use the features that were used by~\citet{gasse2019exact} and~\citet{confs/aaai/KhalilBND16}. 
The list of features in $\mathbf{G}$ are described in the Table~\ref{tab:features_G}. 
There are a total of 92 features that we use for $\mathbf{X}$ as described in the Table~\ref{tab:features_X}.
We follow the preprocessing procedure as described in~\citet{gasse2019exact}.
\input{inputs/features_G}
\input{inputs/features_X}

\section{Preliminary results on running times of various architectures}
In order to have a rough idea of relative improvement in runtimes across various architectures, we considered 20 instances in each difficulty level for each problem class and use the solver to obtain observations at each node. 
Different functions are then used to evaluate decisions at each node, and the total time taken for all the function evaluations across the nodes in an instance is observed. 
Note that the quality of decision is not important here; we are only interested in time taken per decision.

Specifically, we considered 5 forms of architectures as listed in Table~\ref{tab:prelim_models}. 
To cover the entire spectrum of strong inductive biases, we constructed an attention mechanism as explained in~\ref{sec:attn}.
At the same time, to cover the spectrum of cheaper architectures we consider simple dot product as the form of predictor. 
Finally, we consider a scenario where we only consider MLPs as predictors everywhere in the \bnb tree. 
\input{inputs/prelim_models}
\input{inputs/cpu_time_performance}

Figure~\ref{fig:cpu_time} shows the relative performance (rel. to GNN) of various deep learning architectures across 4 sets of problems.
It is evident that MLP ALL and GNN DOT are favored across the problem sets. 
This observation inspired the range of \textit{hybrid architectures} that we explored in the paper.
We also observe that a superior inductive bias like that of attention mechanism and transformers~\cite{vaswani2017attention} has a better runtime performance on GPUs as illustrated in Figure~\ref{fig:gpu_time}, which we suspect is massive parallelization employed in the computations of attention.
However, their performance on CPUs is not better than GNNs. 
\input{inputs/gpu_time_performance}

\section{Attention Mechanism for MILPs}
\label{sec:attn}
To cover the entire spectrum of computational complexity and expressivity of inductive bias, we implemented a transformer~\cite{vaswani2017attention} as an architecture to replace GNNs. 
Specifically, we let the variables (constraints) attend to all other variables (constraints) via multi-headed self-attention mechanism.
Finally, a modulated attention mechanism between \textit{variable representations} as queries and \textit{constraint representation} as keys outputs final \textit{variable representations}, which is passed through the softmax layer for classification objective. 
Here, we use modulation scheme as explained in~\citet{shaw2018self}.
Precisely, an edge in variable-constraint graph is used to increment the attention score with a learnable scalar value. 
\input{inputs/ilp-transformer}

\section{Data Generation \& Training Specifications}
For our experiments, we used 10,000 training instances for the training dataset and collected 150,000 observations from the tree nodes of those instances.
In a similar manner, we generated 20,000 instances each for the validation and testing set, resulting in 30,000 observations each for validation and testing respectively.  

We held all the training parameters fixed across the models.
Specifically, we used a learning rate of $1e^{-3}$, training batch size of $32$, a learning rate scheduler to reduce learning rate by $0.2$ if there is no improvement in the validation loss for $15$ epochs, and an early stopping criterion of $30$ epochs. 
Our epoch consisted of 10K training examples and 2K validation samples. 
We used a 3 layered MLP with 256 hidden units in each layer, while we used GNN model with an embedding size of 64 units.
We used this configuration across all the models discussed in the main paper.
Due to the large size of instances in capacitated facility location, we used a learning rate of 0.005, early stopping criterion of 20 epochs with a patience of 10 epochs.

Further, for knowledge distillation we used $T=2$ (temperature) and $\alpha = 0.9$ (convex mixing of soft and hard objectives). These are the recommended settings in ~\citet{hinton2015distilling}.
We did a hyperparameter search for $\beta = \{0.01, 0.001, 0.0001\}$ for ED and MHE.
Following are the values for $\beta$ that resulted in the best performing models. 

\begin{table}[h]
  \small
  \caption{Best AT models}
  \centering
  \begin{tabular}{ p{4cm} | c  }
    \toprule
    {} & AT \\
    \midrule
    Capacitated Facility Location & MHE \\
    Combinatorial Auctions & MHE \\
    Set Covering & ED \\
    Maximum Independent Set & MHE \\
    \bottomrule
  \end{tabular}
  \label{tabl:arch-types}
\end{table}

We implemented all the models using PyTorch~\citep{paszke2017automatic}, and ran all the CPU evaluations on an Intel(R) Xeon(R) CPU E5-2650 v4 @ 2.20GHz.
GPU evaluations for GNN were performed on NVIDIA-TITAN Xp GPU card with CUDA 10.1. 

\section{Depth-dependent loss weighting scheme}
In Figure~\ref{fig:weight_generalization} we plot the sorted ratio of number of nodes to the minimum number of nodes observed for an instance across all the weighting schemes.
Here we attempt to breakdown the performance of different branching strategies learned using different loss-weighting schemes. 
We observe that the \textit{sigmoidal} scheme achieves the best performance.
\input{inputs/weight_generalization}

\section{Model Results}
The table below is a list of all the architectures along with their performance on the test sets. 
Please note that training with auxiliary task is not possible with HyperSVM type of architecture so their results are not reported in the table. 
\input{inputs/test_results}

\section{Overfitting in maximum independent set}
Table~\ref{tab:overfitting} shows that the FiLM models overfit on small instances of maximum independent set.
To address this problem, we used a mini dataset of 2000 observations obtained by running data collection on medium instances of maximum independent set. 
Further, we regularized the FiLM parameters of the FiLM model to yield much simpler models based on the performance on this mini-dataset, which is not too expensive to obtain owing to the size of observations.
The results of the regularized models are in Table~\ref{tabl:l2-reg-film}
For a fair comparison, we regularized GNN and used the best performing model to report evaluation results in the main paper. 

\begin{table}[htb]
\centering
\caption{FiLM models for maximum independent set overfits on small instances}
    \begin{tabular}{p{20mm}|ccc |ccc|ccc}
    \toprule
    {} & \multicolumn{3}{c}{small} & \multicolumn{3}{c}{medium} & \multicolumn{3}{c}{big} \\
    {} &       Time &     Wins &    Nodes &       Time &     Wins &    Nodes &       Time &     Wins &    Nodes \\
    \midrule
    FiLM &  52.96 &   \textbf{39}/ 55 &    492 &  1515.19 &    9/ 17 &   2804 &  2700.02 &    0/  0 &    nan \\
    GNN-CPU &   \textbf{44.07} &   21/ 60 &    \textbf{432} &   \textbf{371.81} &   \textbf{36}/ 40 &    \textbf{558} &  \textbf{1981.43} &   \textbf{10}/ 10 &   \textbf{6334}\\
    \midrule
    GNN-GPU  & 31.70 &    0/ 59 &    432 &   264.01 &    0/ 43 &    558 &  1772.12 &    0/ 13 &   6313 \\
    \bottomrule
    \end{tabular}
    \label{tab:overfitting}
\end{table}

\begin{table}[h!]
  \small
  \caption{Top-1 accuracy of regularized models on 2000 observations from medium random instances of maximum independent set.}
  \centering
  \begin{tabular}{ p{2cm} | c | c  }
    \toprule
    weight decay & FiLM & GNN \\
    \midrule
    1.0 & 55.15 $\pm$ 0.07 & 31.6 $\pm$ 6.63 \\
    0.1 & \textbf{56.13 $\pm$ 0.32} & \textbf{37.23 $\pm$ 0.92} \\
    0.01 & 53.25 $\pm$ 0.95 & 26.8 $\pm$ 16.71 \\
    0.0 & 19.18 $\pm$ 4.24 & 34.08 $\pm$ 4.8 \\
    \bottomrule
  \end{tabular}
  \label{tabl:l2-reg-film}
\end{table}

\section{Performance of HyperSVM architectures}
HyperSVM architectures are the cheapest in computation. 
In this section we ask if we are willing to lose machine learning accuracy by 1-2\% in HyperSVM architecture, can we still get faster running times with HyperSVM type of architectures?
Table~\ref{tab:hypersvm-evaluation} compares solver performance by using HyperSVM architecture and FiLM architecture. 
We observe that HyperSVM types of architectures loose their ability to generalize on larger scale instances. 
\input{inputs/hypersvm-comparison}

\section{Scaling to twice the size of Big instances}
In this section we investigate the generalization power of FiLM models trained on dataset obtained from small instances. 
Specifically, we generate 20 random instances of size double that of big instances, and use the trained models of FiLM and GNN to compare their performance against RPB. 
We use the time limit of 7200 seconds, i.e. 2 hours to account for longer solving running times as one scales out to bigger instances.
We observe that the power to generalize is largely dependent on the problem family.
For example, FiLM models can still outperform other strategies on scaling out on capacitated facility location problems. 
However, we found that RPB remains competitive on setcover problems.
We note that this shortcoming of the FiLM models can be overcome via larger size of hidden layers and training on slightly larger instances than what has been used in the main paper.
Table~\ref{tab:scale-out} shows these results. 
\input{inputs/scale-out}

\section{Effect of different training protocols on B\&B performance}
Table~\ref{tab:protocol-evaluation} shows the performance of different training protocols on \bnb{} performance. 
Although the models trained with auxiliary tasks (AT) and knowledge distillation (KD) are a clear winner up until problem sets of size medium, there is a tie between the models trained with KD and those trained with KD \& AT.
However, it is worth noting that the difference in \bnb{} performance across different training protocols is not huge, and such performance evaluation can be read from the test accuracy of trained models. 
For example, as evident in Table~\ref{tab:test-perf}, difference in test accuracy of FiLM (e2e \& KD) and FiLM (e2e \& KD \& AT) is not significant for problem sets - Capacitated Facility Location and Set Covering, but its significant enough for problem sets - Combinatorial Auctions and Maximum Independent Set. 
\textbf{Given that the inference cost is independent of training protocols, to attain better generalization performance, we recommend FiLM (e2e \& KD \& AT) only when these models have significantly better accuracy than FiLM (e2e \& KD).}
\input{inputs/training-protocol-performance}

\small
\bibliography{references}

%% file: inputs/times_arch.tex
\begin{wrapfigure}{r}{0.45\textwidth}
    \vspace{0pt}
    \centering
    \includegraphics[clip, trim=0 0 0 0cm, width=\linewidth]{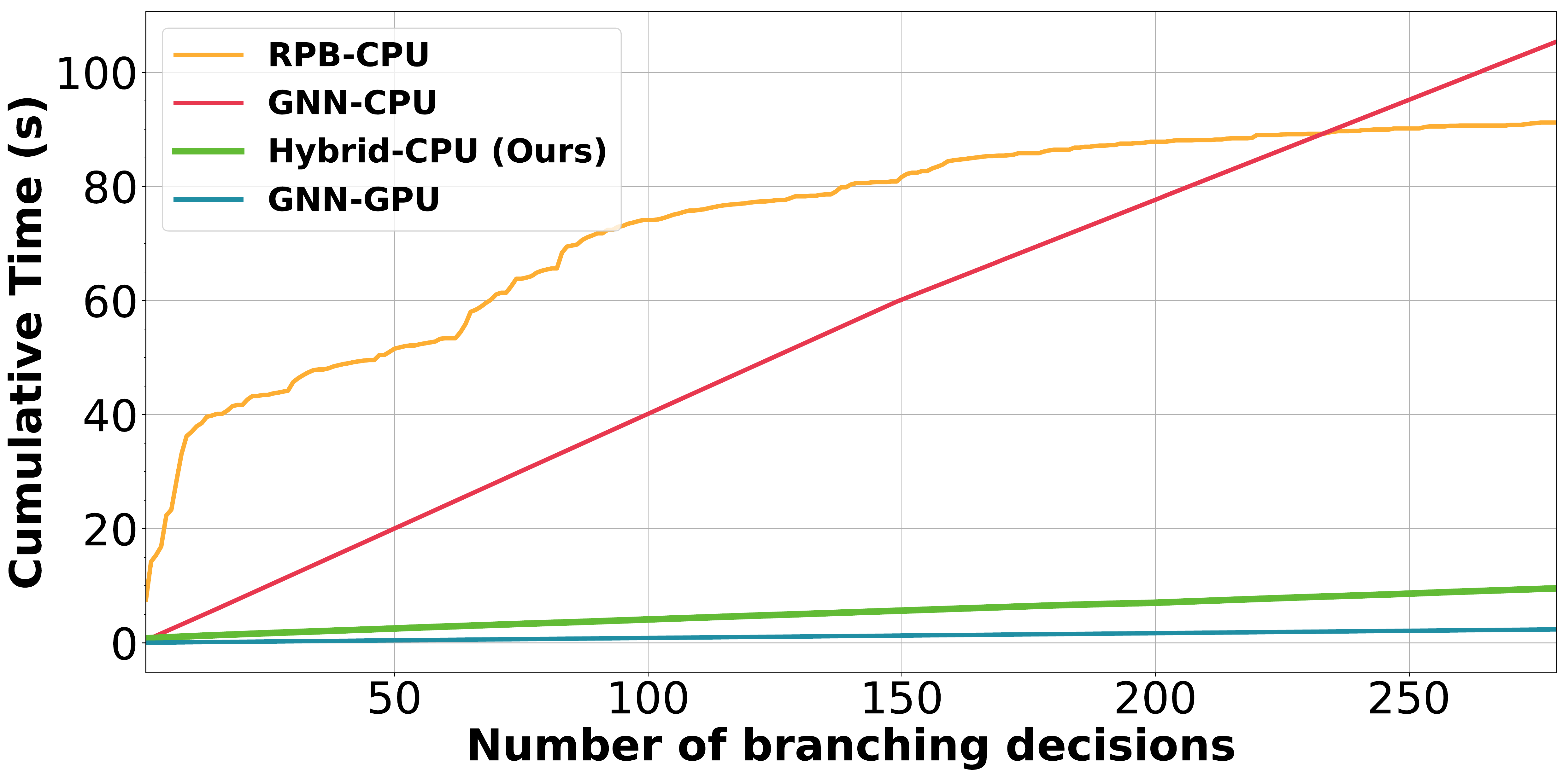}
    \caption{Cumulative time cost of different branching policies: (i) the default internal rule RPB of the SCIP solver; (ii) a GNN model (using a GPU or a CPU); and (iii) our hybrid model. Clearly the GNN model requires a GPU for being competitive, while our hybrid model does not. (Measured on a capacitated facility location problem, medium size).}
    \label{fig:times_arch}
    \vspace{0pt}
\end{wrapfigure}

%% file: inputs/arch-data-types.tex
\begin{wrapfigure}{r}{0.48\textwidth}
    \vspace{-1mm}
    \centering
    \includegraphics[width=0.5\textwidth]{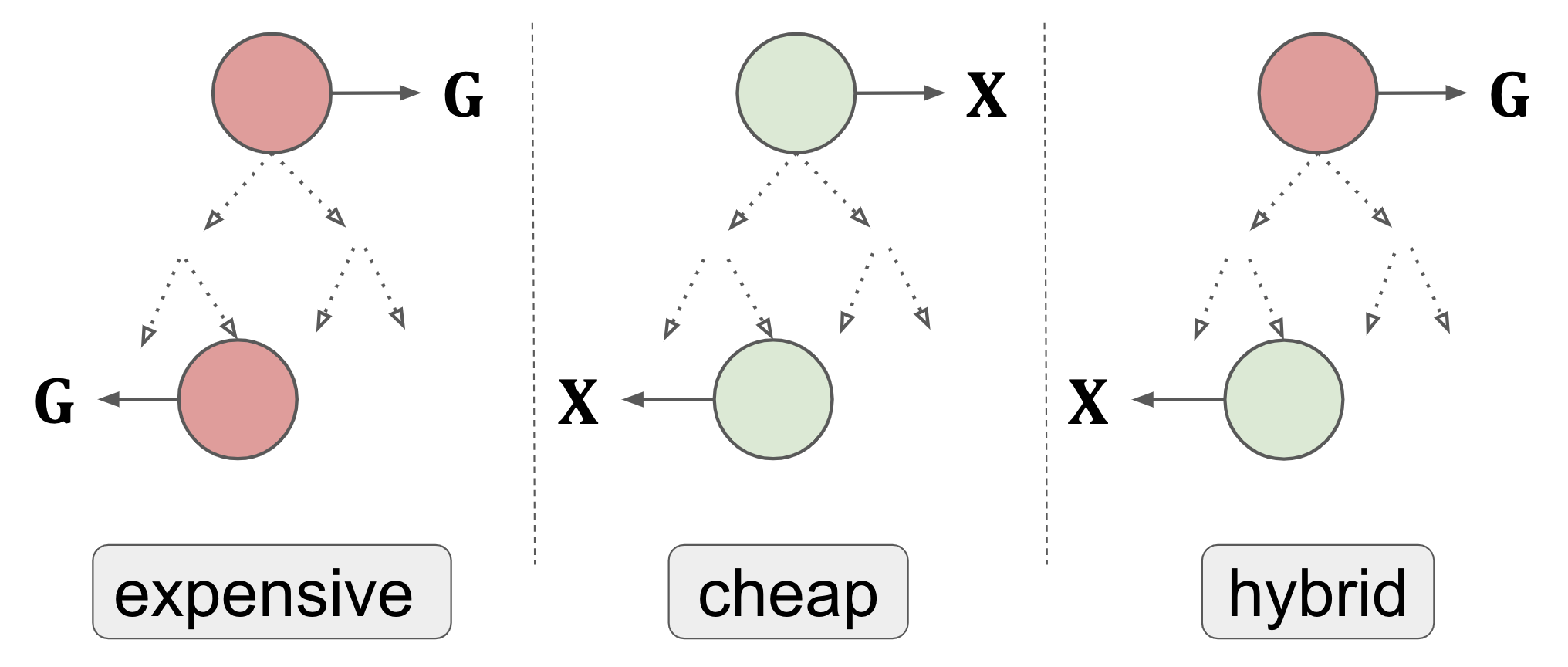}
    \caption{Data extraction strategies: bipartite graph representation $\mathbf{G}$ at every node (expensive); candidate variable features $\mathbf{X}$ at every node (cheap); bipartite graph at the root node and variable features at tree node (hybrid).}
    \label{fig:arch-data-type}
    \vspace{-5mm}
\end{wrapfigure}

%% file: inputs/arch-types.tex
\begin{table}[t]
  \setlength\extrarowheight{-3pt}
  \captionsetup{justification=centering}
  \small
  \caption{Various functional forms $f$ considered for variable selection ($\odot$ denotes Hadamard product).}
  \renewcommand{\arraystretch}{1.3}
  \centering
  \begin{tabular}{ p{2cm} | c | c | p{5.5cm} }
    \toprule
     {} &
     Data extraction &
     Computational Cost &
     \multicolumn{1}{c}{Decision Function}  \\
    \midrule
     GNN~\cite{gasse2019exact} & expensive & Expensive & $\begin{array} {lcl} \mathbf{s} & = & \text{GNN}(\mathbf{G})\end{array}$  \\ 
    \midrule
    MLP & cheap & Moderate & $\begin{array} {lcl} \mathbf{s} & = & \text{MLP}(\mathbf{X})\end{array}$   \\ 
    \midrule
    CONCAT & hybrid & Moderate & $\begin{array} {lcl} \pmb{\Psi} & = & \text{GNN}(\mathbf{G}^0) \\ \mathbf{s} & = & \text{MLP}([\pmb{\Psi}, \mathbf{X}]) \end{array}$  \\
    \midrule 
    FiLM~\citep{perez2018film}  & hybrid & Moderate & $\begin{array} {lcl} \pmb{\gamma}, \pmb{\beta} & = & \text{GNN}(\mathbf{G}^0) \\ \mathbf{s}  & = & \text{FiLM}(\pmb{\gamma}, \pmb{\beta}, \text{MLP}(\mathbf{X})) \end{array}$ \\
    \midrule 
    HyperSVM & hybrid & Cheapest & $\begin{array} {lcl} \mathbf{W} & = & \text{GNN}(\mathbf{G}^0) \\ \mathbf{s} & = & (\mathbf{W} \odot \mathbf{X})\mathbf{1} \end{array}$ \\
    \midrule
     HyperSVM-FiLM & hybrid & Cheapest & $\begin{array} {lcl} \pmb{\gamma}, \pmb{\beta_1}, \pmb{\beta_2} & = & \text{GNN}(\mathbf{G}^0) \\ \mathbf{s} & = & \pmb{\beta_2}^\intercal \max(0, \pmb{\beta_1} \odot \mathbf{X} + \pmb{\gamma})  \end{array}$ \\
    \bottomrule
  \end{tabular}
  \label{tabl:arch-types}
  \vspace{-15pt}
\end{table}

%% file: inputs/model_results.tex
\begin{wrapfigure}{r}{0.60\textwidth}
    \vspace{-4mm}
    \begin{center}
    \includegraphics[width=0.6\textwidth]{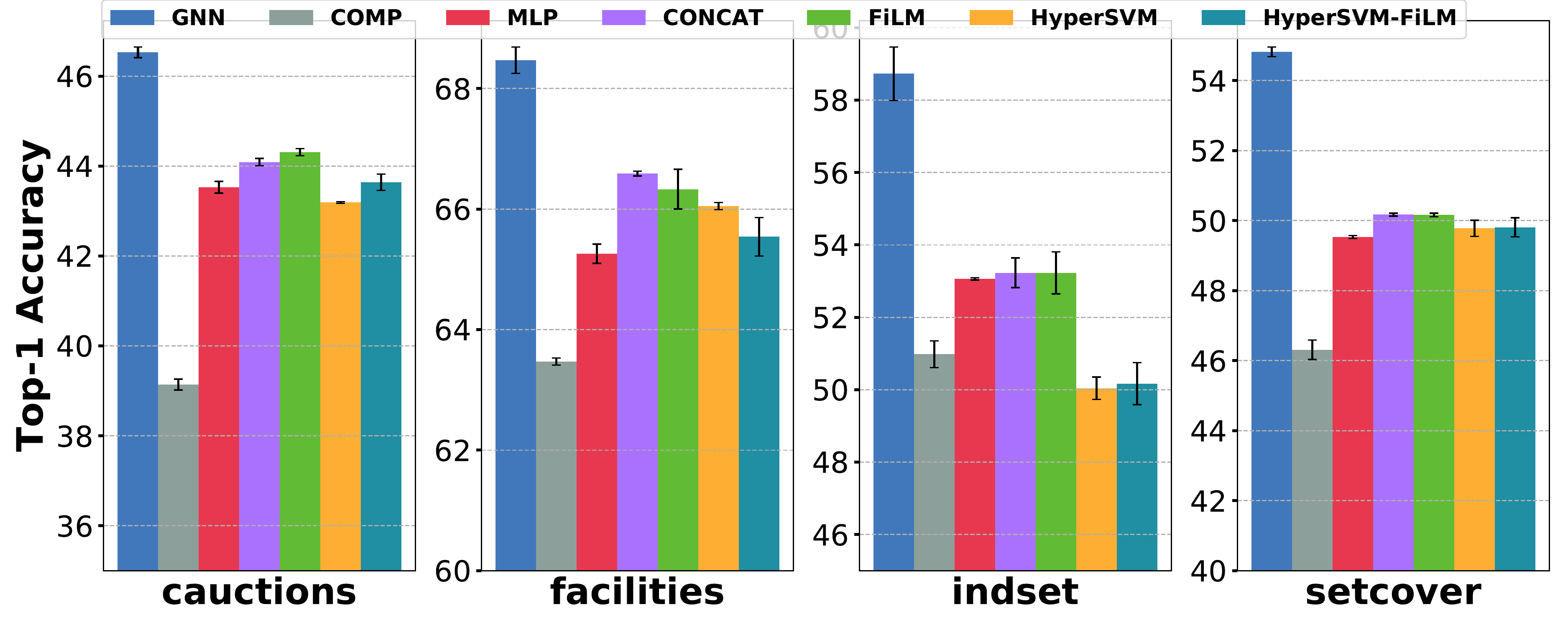}
    \end{center}
       \caption{Test accuracy of the different models, with a simple e2e training protocol. 
    }
    \label{fig:model_results}
    \vspace{-2mm}
\end{wrapfigure}

%% file: inputs/combined-protocol-loss-weighting.tex
\begin{wraptable}{r}{0.65\textwidth}
\vspace{-5pt}

  \scriptsize
  \caption{Test accuracy of FiLM, using different training protocols.}
  \centering
  \setlength\extrarowheight{-3pt}
  \begin{tabular}{lllll}
     {}
     & \multicolumn{1}{c}{cauctions}
     & \multicolumn{1}{c}{facilities}
     & \multicolumn{1}{c}{indset}
     & \multicolumn{1}{c}{setcover}
     \\
    \toprule
    Pretrained GNN & 44.12 $\pm$ 0.09 & 65.78 $\pm$ 0.06  & 53.16 $\pm$ 0.51 & 50.00 $\pm$ 0.09 \\ 
    e2e  & 44.31 $\pm$ 0.08 & 66.33 $\pm$ 0.33 & 53.23 $\pm$ 0.58 & 50.16 $\pm$ 0.05 \\ 
    e2e \& KD & 44.10 $\pm$ 0.09 & 66.60 $\pm$ 0.21 & 53.08 $\pm$ 0.3 & 50.31 $\pm$ 0.19 \\
    e2e \& KD \& AT & \textbf{44.56 $\pm$ 0.13} & \textbf{66.85 $\pm$  0.28} & \textbf{53.68 $\pm$ 0.23} & \textbf{50.37 $\pm$ 0.03} \\
    \bottomrule
  \end{tabular}
  \label{tabl:protocol-results}

    \vspace{10pt}

    \setlength\extrarowheight{-3pt}
    \scriptsize
    \caption{Effect of different sample weighting schemes on combinatorial auctions (big) instances, with a simple MLP model. $z \in [0,1]$ is the ratio of the depth of the node and the maximum depth observed in a tree.}
    \centering
    \begin{tabular}{llrr}
    \multicolumn{1}{c}{Type} &
    \multicolumn{1}{c}{Weighting scheme} &
    Nodes &
    Wins \\
    \toprule
    Constant & $1$ & 9678 & 10/60\\ 
    Exponential decay & $e^{-0.5z}$ & 9793 & 10/60\\
    Linear & $(e^{-0.5} - 1)*z + 1$ & 9789 & 12/60\\
    Quadratic decay & $(e^{-0.5} - 1)*z^2 + 1$ & 9561 & 14/60 \\
    Sigmoidal & $(1+e^{-0.5})/(1+e^{z-0.5}) $ & \textbf{9534} & 14/60 \\
    \bottomrule
    \end{tabular}
    \label{tab:weights}
\vspace{-5pt}
\end{wraptable}

%% file: inputs/evaluation-compressed-format.tex
\sisetup{detect-weight=true,detect-inline-weight=math,detect-mode=true}
\begin{table}[t]
\caption{
Performance of \textit{branching strategies} on evaluation instances. We report geometric mean of solving times, number of times a method won (in solving time) over total finished runs, and geometric mean of number of nodes. Refer to section~\ref{sec:experiments} for more details. The best performing results are in \textbf{bold}. $^*$Models were regularized to prevent overfitting on small instances.}
\label{tab:evaluation}
\centering
\scriptsize
\setlength{\tabcolsep}{10pt}
\aboverulesep = 0.1mm  
\belowrulesep = 0.2mm  
\begin{tabular}{
    c
    S[table-format=3.2]@{\hspace{10pt}}
    S[table-format=2.0]@{\hspace{4pt}/\hspace{-4pt}}
    S[table-format=2.0]@{\hspace{5pt}}
    S[table-format=3.0]@{}
    S[table-format=3.2]@{\hspace{10pt}}
    S[table-format=2.0]@{\hspace{4pt}/\hspace{-4pt}}
    S[table-format=2.0]@{\hspace{5pt}}
    S[table-format=3.0]@{}
    S[table-format=3.2]@{\hspace{10pt}}
    S[table-format=2.0]@{\hspace{4pt}/\hspace{-4pt}}
    S[table-format=2.0]@{\hspace{5pt}}
    S[table-format=3.0]@{}
    }
    \toprule
     &
    \multicolumn{4}{ c }{Small} &
    \multicolumn{4}{ c }{Medium} &
    \multicolumn{4}{ c }{Big} \\

    Model &
    \multicolumn{1}{ c }{Time} &
    \multicolumn{2}{ c }{Wins} &
    \multicolumn{1}{ c }{Nodes} &
    \multicolumn{1}{ c }{Time} &
    \multicolumn{2}{ c }{Wins} &
    \multicolumn{1}{ c }{Nodes} &
    \multicolumn{1}{ c }{Time} &
    \multicolumn{2}{ c }{Wins} &
    \multicolumn{1}{ c }{Nodes} \\
    \toprule


    \textsc{fsb}
        &   42.53 &   1 &  60 &   13
        &  313.33 &   0 &  59 &   75
        &  997.23 &   0 &  51 &   50  \\

    \cmidrule(lr){1-5} \cmidrule(lr){6-9} \cmidrule(lr){10-13}

    \textsc{pb}
        &   31.35 &   4 &  60 &  139
        &  177.69 &   4 &  60 &  384
        &  712.45 &   3 &  56 &  309  \\

    \textsc{rpb}
        &   36.86 &   1 &  60 &  \B  23
        &  213.99 &   1 &  60 &  \B 152
        &  794.80 &   2 &  54 &  \B 99 \\

    \textsc{comp}
        &   30.37 &   3 &  60 & 120 
        &  172.51 &   4 &  60 & 347 
        &  633.42 &   6 &  57 & 294  \\

    \textsc{gnn}
        &   39.18 &   0 & 60 &  112 
        &  209.84 &   0 & 60 &  314 
        &  748.85 &   0 & 54 &  286 \\

    \textsc{FiLM} (ours)
        & \B   24.67 & \B  51 &  60 &  109 
        & \B  136.42 & \B  51 &  60 &  325
        & \B  531.70 & \B  46 &  57 &  295  \\

    \cmidrule(lr){1-5} \cmidrule(lr){6-9} \cmidrule(lr){10-13}
    \textsc{gnn-gpu}
        &      28.91 &     \text{--} &  60 &  112 
        &     150.11 &     \text{--} &  60 &  314
        &     628.12 &     \text{--} &  56 &  286  \\
        
    \cmidrule(lr){1-5} \cmidrule(lr){6-9} \cmidrule(lr){10-13}
    \\[-7pt]
    & \multicolumn{12}{ c }{Capacitated Facility Location} \\
    \\[-3pt]

    \cmidrule(lr){1-5} \cmidrule(lr){6-9} \cmidrule(lr){10-13}

    \textsc{fsb}
        &   27.16 &    0 &  60 & 17
        &  582.18 &    0 &  45 & 116 
        & 2700.00 &    0 &   0 & n/a  \\

    \cmidrule(lr){1-5} \cmidrule(lr){6-9} \cmidrule(lr){10-13}
    \textsc{pb}
        &   10.19 &   0 &  60 & 286
        &   94.12 &   0 &  60 & 2451
        & 2208.57 &   0 &  23 & 82624  \\

    \textsc{rpb}
        &   14.05 &    0 &  60 & \B     54
        &   94.65 &    0 &  60 & 1129
        & 1887.70 &    7 &  27 & 48395 \\

    \textsc{comp}
        &    9.83 &   3 &  60 & 178
        &   89.24 &   0 &  60 & 1474
        & 2166.44 &   0 &  21 & 52326  \\

    \textsc{gnn}
        &   17.61 &   0 &  60 & 136
        &  242.15 &   0 &  60 & \B  1013
        & 2700.17 &   0 &   0 & n/a \\

    \textsc{FiLM} (ours)
        & \B    8.73 & \B 57 &     60 & 147
        & \B   63.75 & \B 60 &     60 & 1131
        & \B 1843.24 & \B 20 &     26 & \B  37777 \\

    \cmidrule(lr){1-5} \cmidrule(lr){6-9} \cmidrule(lr){10-13}
    \textsc{gnn-gpu}
        &       8.26 &     \text{--} &  60 &  136
        &      53.56 &     \text{--} &  60 &  1013
        &    1535.80 &     \text{--} &  36 &  31662  \\
    \cmidrule(lr){1-5} \cmidrule(lr){6-9} \cmidrule(lr){10-13}

    \\[-7pt]
    & \multicolumn{12}{ c }{Set Covering} \\
    \\[-3pt]

    \cmidrule(lr){1-5} \cmidrule(lr){6-9} \cmidrule(lr){10-13}

    \textsc{fsb}
        &    6.12 &    0 &  60 & 6
        &  132.38 &    0 &  60 & 71
        & 2127.35 &    0 &  28 & 318  \\

    \cmidrule(lr){1-5} \cmidrule(lr){6-9} \cmidrule(lr){10-13}
    \textsc{pb}
        &    2.76 &   1 &  60 & 234
        &   25.83 &   0 &  60 & 2765
        &  393.60 &   0 &  59 & 13719  \\

    \textsc{rpb}
        &    4.01 &       0 &  60 & \B 11
        &   26.36 &       0 &  60 & 714
        &  \B 210.95 & \B   29 &  60 & 4701 \\

    \textsc{comp}
        &    2.76 &   0 &  60 & 82
        &   29.76 &   0 &  60 & 930
        &  494.59 &   0 &  54 & 5613  \\

    \textsc{gnn}
        &    2.73 &   1 &  60 & 71
        &   22.26 &   0 &  60 & 688
        &  257.99 &   6 &  60 & \B 3755 \\

    \textsc{FiLM} (ours)
        & \B    2.13 & \B 58 &     60 & 73
        & \B   15.71 & \B 60 &     60 &\B  686
        &     217.02 &    25 &     60 &  4315 \\

    \cmidrule(lr){1-5} \cmidrule(lr){6-9} \cmidrule(lr){10-13}
    \textsc{gnn-gpu}
        &       1.96 &     \text{--} &  60 &  71
        &      11.70 &     \text{--} &  60 &  688
        &     121.18 &     \text{--} &  60 &  3755  \\
    \cmidrule(lr){1-5} \cmidrule(lr){6-9} \cmidrule(lr){10-13}

    \\[-7pt]
    & \multicolumn{12}{ c }{Combinatorial Auctions} \\
    \\[-3pt]

    \cmidrule(lr){1-5} \cmidrule(lr){6-9} \cmidrule(lr){10-13}

    \textsc{fsb}
        &  673.43 &    0 &  53 & 47
        & 1689.75 &    0 &  20 & 10
        & 2700.00 &    0 &   0 & n/a  \\

    \cmidrule(lr){1-5} \cmidrule(lr){6-9} \cmidrule(lr){10-13}
    \textsc{pb}
        &  172.03 &   2 &  57 & 5728
        &  753.95 &   0 &  45 & 1570
        & 2685.23 &   0 &   1 & 38215  \\

    \textsc{rpb}
        &   59.87 &       5 &  60 & 603
        &  173.17 &      11 &  60 & \B 205
        & 1946.51 &       9 &  21 & 2461 \\

    \textsc{comp}
        &   82.22 &  1 &  58 & 847
        &  383.97 &  1 &  52 & 267
        & 2393.75 &  0 &   6 & 5589  \\

    \textsc{gnn$^\star$}
        &\B 44.07 &  15 &  60 & \B 331
        &  625.23 &   1 &  50 & 599
        & 2330.95 &   0 &  10 & \B 687 \\

    \text{FiLM$^\star$} (ours)
        &      52.96 & \B 37 &     55 & 376
        & \B  131.45 & \B 47 &     54 & 264
        & \B 1823.29 & \B 12 &     15 &  1201 \\

    \cmidrule(lr){1-5} \cmidrule(lr){6-9} \cmidrule(lr){10-13}
    \textsc{gnn-gpu$^\star$}
        &      31.71 &     \text{--} &  60 &  331
        &      63.96 &     \text{--} &  60 &  599
        &    1158.59 &     \text{--} &  27 &  685  \\
        
    \cmidrule(lr){1-5} \cmidrule(lr){6-9} \cmidrule(lr){10-13}
    \\[-7pt]
     & \multicolumn{12}{ c }{Maximum Independent Set} \\
     \bottomrule
\end{tabular}
\vspace{-15pt}
\end{table}

%% file: inputs/optimality-gap.tex
\begin{wraptable}{r}{0.35\textwidth}
      \vspace{-5pt}
      \captionsetup{justification=centering}
      \scriptsize
      \caption{Mean optimality gap (lower the better) of commonly unsolved ``big" instances (number of such instances in brackets).}
      \centering
      \setlength\extrarowheight{-3pt}
      \begin{tabular}{lll}
        \toprule
         {}
         & \multicolumn{1}{c}{setcover (33) }
         & \multicolumn{1}{c}{indset (39)}
         \\
        \midrule
        FSB  & 0.1709 & 0.0755  \\
        \midrule
        PB   & 0.0713 & 0.0298 \\
        RPB  & 0.0628 & 0.0252 \\
        COMP & 0.0740 & 0.0252 \\
        GNN  & 0.1039 & 0.0341 \\
        FiLM & \textbf{0.0597} & \textbf{0.0187} \\
        \bottomrule
      \end{tabular}
      \label{table:opt-gap}
      \vspace{-5pt}
\end{wraptable}

%% file: inputs/batch_pack.tex
\begin{figure}[htb]
    \centering
    \includegraphics[width=0.7\textwidth]{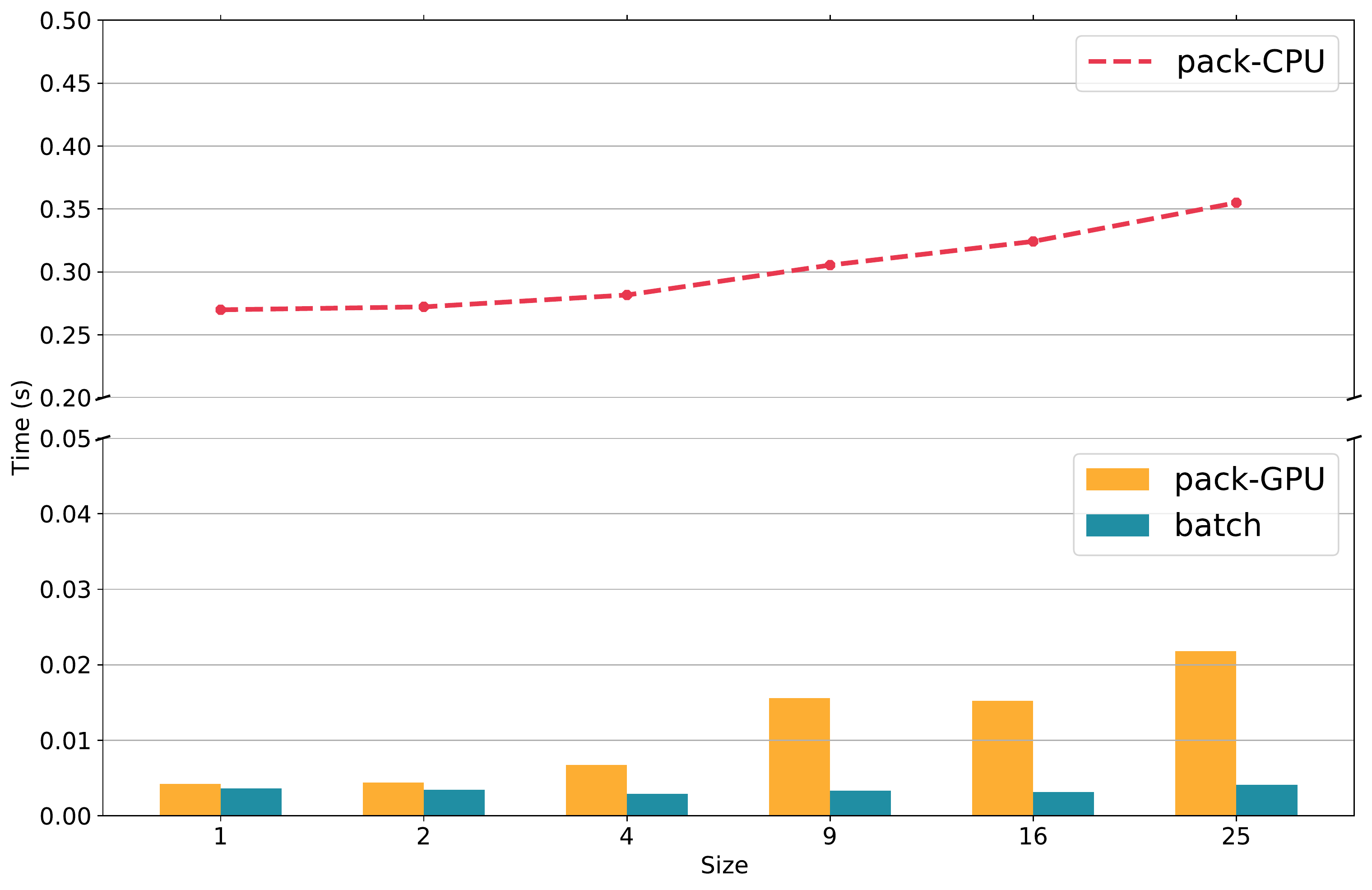}
    \caption{\textit{Packing} several GNNs together on a GPU keeps it underutilized. ``Size'' is the number of inputs batched together (unrealistic scenario) or number of inputs simultaneously put on a GPU separately.}
    \label{fig:batch_pack}
\end{figure}

%% file: inputs/features_G.tex
\begin{table}[htb]
    \centering
    \setlength\extrarowheight{-3pt}
    \captionsetup{justification=centering}
    \small
    \caption{Features of $\mathbf{G}$. It is constructed as a variable-constraint bipartite graph, where the vertices of this graph have features as described here. These features are same as used in~\citet{gasse2019exact}.}
    \label{tab:features_G}
    \begin{tabular}{l l p{7cm}}
    \toprule 
     {} & Type & Description \\
    \midrule
    Variable Features (Total -13) & & \\ 
    \midrule
    Variable type (1) & Categorical & Allowed values -  Binary, Integer, Continuous, Implied Integer \\
    Normalized coefficient (1) & Real & Objective coefficient of the variable normalized by the euclidean norm of its coefficients in the constraints  \\
    Specified bounds (2) & Binary & Does the variable has a lower bound (upper bound)? \\ 
    Solution bounds (2) & Binary & If the variable is currently at its lower bound (upper bound)? \\ 
    Solution bractionality (1) & Real $\in [0,1)$ & Fractional part of the variable i.e. $x- \left \lfloor{x}\right \rfloor $, where $x$ is the value of the decision in variable in the current LP solution \\
    Basis (1) & Categorical & One of 4 classes - Lower (variable is at the lower bound), Basic (variable has a value between the bounds), Upper (variable is at the upper bound), Zero (rare case) \\
    Reduced cost (1) & Real & Amount by which objective coefficient of the variable should decrease so that the variable assumes a positive value in the LP solution \\
    Age (1) & Real & Number of LP iterations since the last time the variable was basic normalized by total number of LP iterations \\
    Solution value (1) & Real & Value of the variable in the current LP solution \\ 
    Primal value (1) & Real & Value of the variable in the current best primal solution \\ 
    Average primal value (1) & Real & Average value of the variable in all of the previously observed feasible primal solutions \\ 
    \midrule 
    Constraint Features (Total - 5) \\
    \midrule
    \midrule
    Cossine similarity (1) & Real & Cossine of angle between the vector represented by objective coefficients and the coefficients of this constraint \\
    Bias (1) & Real & Right hand side of the constraint normalized by the euclidean norm of the row coefficients\\
    Age (1) & Real & Number of iterations since the last time the constraint was active normalized by total number of LP iterations  \\
    Normalized dual value (1) & Real & Value of dual variable corresponding to the constraint normalized by the product of norms of the row coefficients and the objective coefficients \\
    Bounds (1) & Binary & If the constraint is currently at its bounds? \\
    \midrule 
    Edge Features (Total - 1)\\ 
    \midrule 
    \midrule
    Normalized coefficient (1) & Real & Coefficient of the variable normalized by the norm of the coefficients of all the variables in the constraint\\
    \bottomrule 
    \end{tabular}
\end{table}

%% file: inputs/features_X.tex
\begin{table}[htb]
    \centering
    \caption{Features in $\mathbf{X}$, an input to MLP.}
    \label{tab:features_X}
    \begin{tabular}{l l p{6.5cm}}
    \toprule 
     {} & count \\
    \midrule
    Variable features from $\mathbf{G}$ & 13  \\ 
    \midrule
    Variable features from~\citet{confs/aaai/KhalilBND16} & 72 \\
    \bottomrule 
    \end{tabular}
\end{table}

%% file: inputs/prelim_models.tex
\begin{table}[htb]
    \scriptsize
    \centering
    \caption{Different architectures considered for preliminray runtime comparison.}
    \begin{tabular}{|l|p{100mm}|}
    \toprule
    Type & Description \\
    \midrule
    GNN ALL & Use a Graph Convolution Neural Network (GNN)~\cite{gasse2019exact} at all \textit{tree nodes}\\
    ATTN ALL & Attention mechanism (section ~\ref{sec:attn}) at all nodes  \\
    GNN DOT & Use GNN at the \textit{root nodes} and a dot product with $\mathbf{X}$ to compute scores at every node \\
    ATTN DOT & Use attention mechanism at the root node and dot product with $\mathbf{X}$ to compute scores at every node \\
    MLP ALL & Use a 3-layer multilinear perceptron at all the nodes\\
    \bottomrule
    \end{tabular}
    \label{tab:prelim_models}
\end{table}

%% file: inputs/cpu_time_performance.tex
\begin{figure}[htb]
\vspace{-2mm}
\begin{center}
\includegraphics[width=0.90\linewidth]{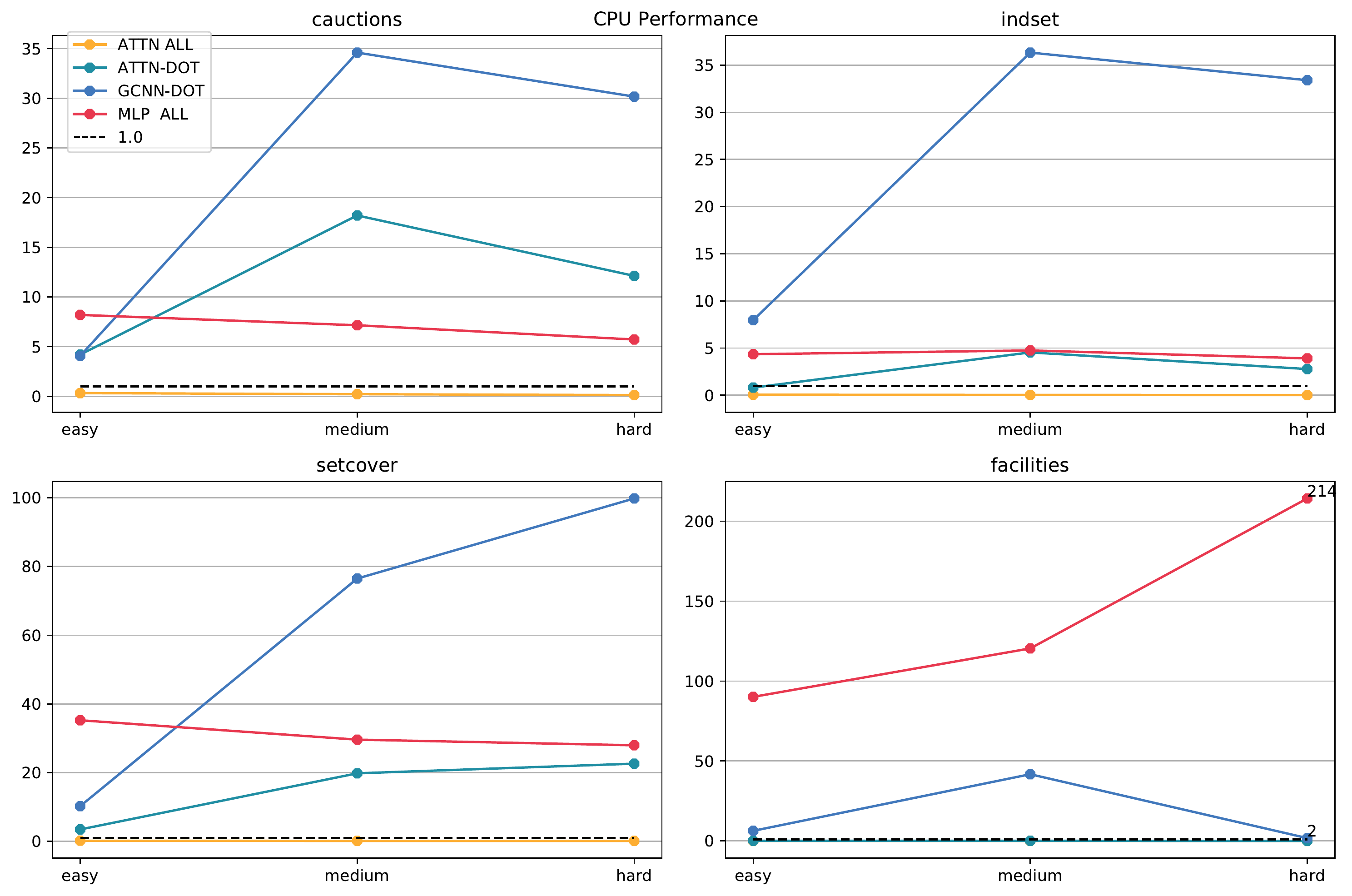}
\end{center}
   \caption{Relative time performance of various methods with respect to GNN ALL on CPU (average values). A value of $10$ implies that the method is 10 times faster (on arithmetic average) than GNN ALL implying that one can afford to perform 10 times worse than GNN ALL in iterative performance when using CPUs. Note that this is just a rough estimation.}
\label{fig:cpu_time}
\end{figure}

%% file: inputs/gpu_time_performance.tex
\begin{figure}[htb]
\vspace{-2mm}
\begin{center}
\includegraphics[width=0.90\linewidth]{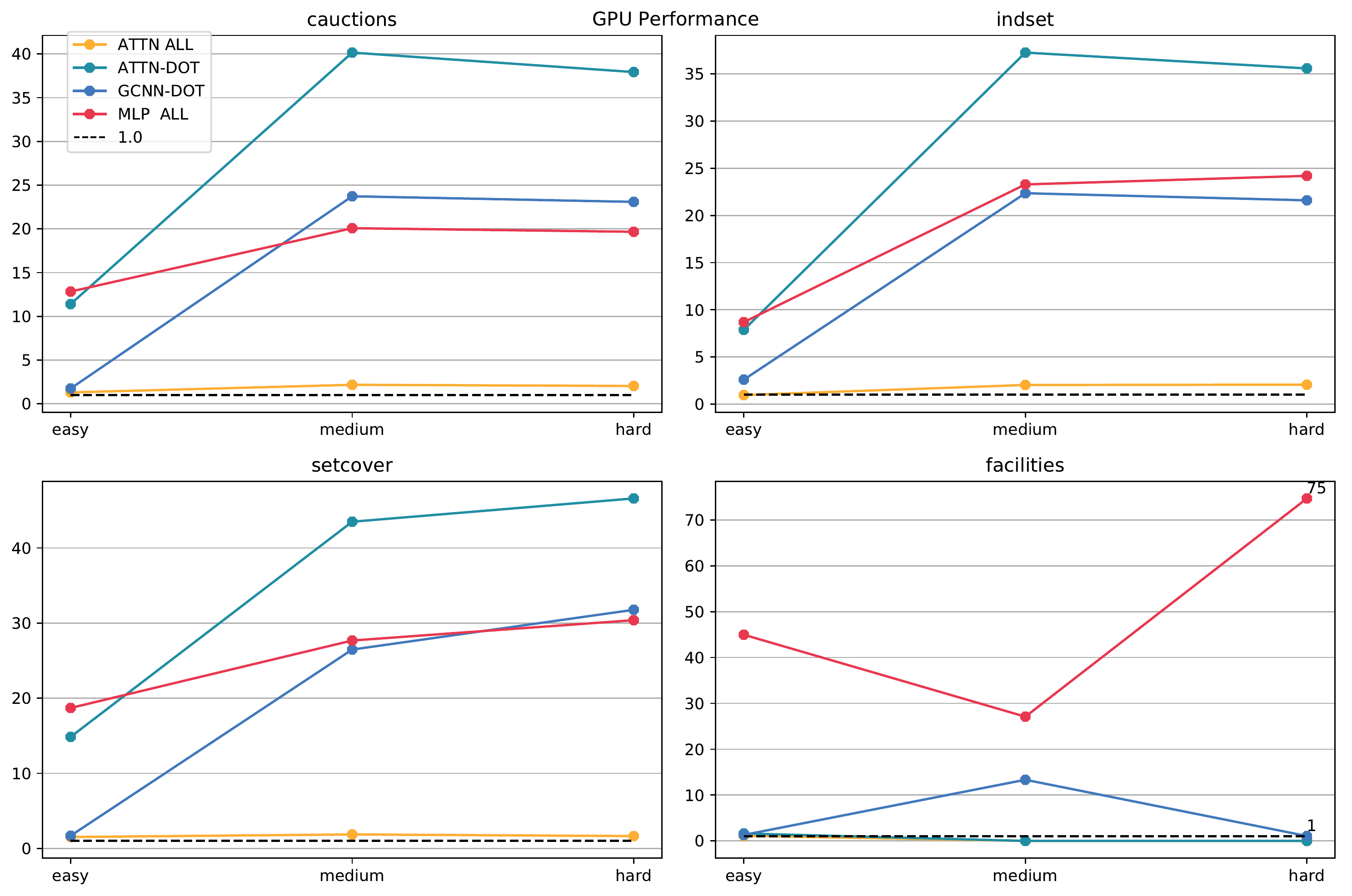}
\end{center}
   \caption{Relative time performance of various methods with respect to $GNN ALL$ on GPU (average values). A value of $10$ implies that the method is 10 times faster (on arithmetic average) than $GNN ALL$ implying that one can afford to perform 10 times worse than $GNN ALL$ in iterative performance when using CPUs. Note that this is just a rough estimation.}
\label{fig:gpu_time}
\vspace{-5mm}
\end{figure}

%% file: inputs/ilp-transformer.tex
\begin{figure}[htb]
\begin{center}
\includegraphics[width=0.7\linewidth]{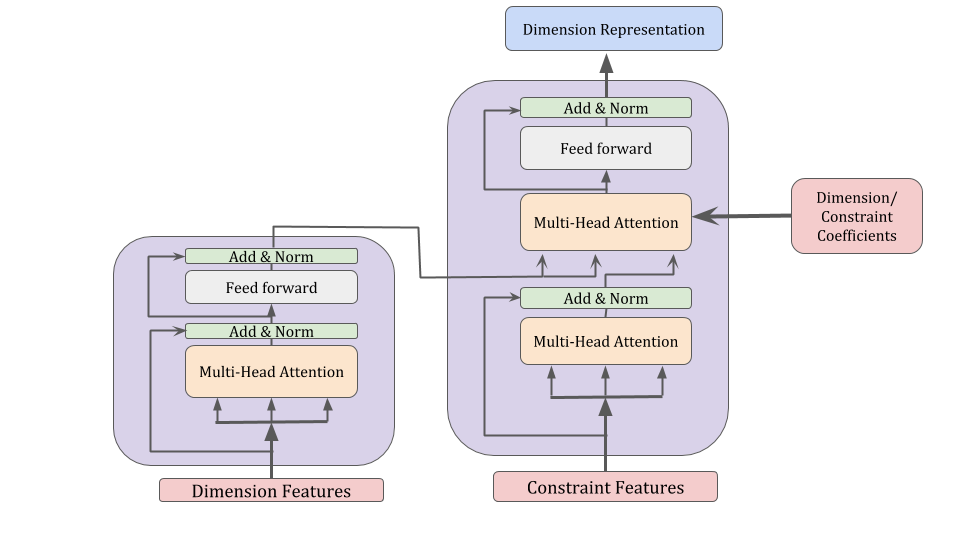}
\end{center}
   \caption{Multi-Head Attention mechanism where a variable or constraint can attend to all other variables or constraints. Finally,variables are used as a query to attend to constraints, where the attention is modulated through variable-constraint features. Modulation of attention scores follow~\citet{shaw2018self}}
\label{fig:ilp-transformer}
\end{figure}

%% file: inputs/weight_generalization.tex
\begin{figure}[htb]
    \centering
    \includegraphics[width=0.7\textwidth]{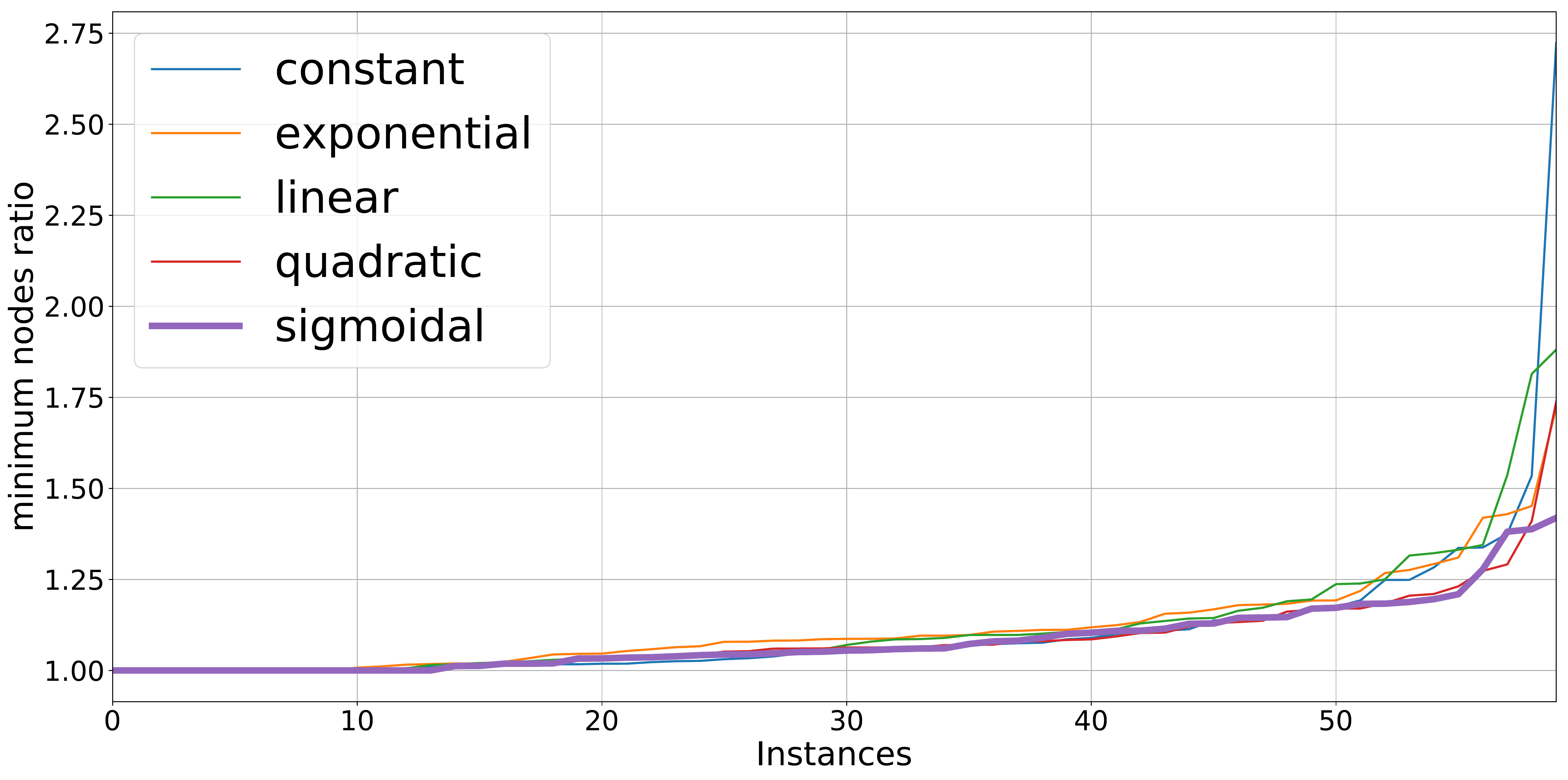}
    \caption{Performance of different weighing schemes across "big" instances. "Sigmoidal" evolves slowest among all.}
    \label{fig:weight_generalization}
\end{figure}

%% file: inputs/test_results.tex
\begin{table}[htb]  
    \scriptsize
    \caption{Top-1 accuracy of various models.}
    \centering
    \begin{tabular}{p{25mm}|p{3cm}|p{1.5cm}p{1.5cm}p{1.5cm}p{1.5cm}}
    \toprule
    {} & {} & Combinatorial Auctions & Capacitated Facility Location & Set Cover & Maximum Independent Set \\
    \midrule
   Expert & GCNN &  46.53 $\pm$ 0.12 & 68.47 $\pm$ 0.22 & 54.82 $\pm$ 0.14 & 58.73 $\pm$ 0.54 \\
   \midrule
   \multirow{3}{*}{Existing methods} & Extratrees &   37.97 $\pm$ 0.1  &  60.06 $\pm$ 0.18 & 42.66 $\pm$ 0.22 & 19.35 $\pm$ 0.55 \\
   & LMART   &   38.7 $\pm$ 0.16 & 63.28 $\pm$ 0.06 & 46.31 $\pm$ 0.28 & 50.98 $\pm$ 0.37  \\
   & SVMRank &  39.14 $\pm$ 0.12 & 63.47 $\pm$ 0.06 & 46.23 $\pm$ 0.11 & 49.29 $\pm$ 0.23  \\
   \midrule
    Simple Network & MLP  &  43.53 $\pm$ 0.13 &  65.26 $\pm$ 0.16 &   49.53 $\pm$ 0.04  & 53.06 $\pm$ 0.03 \\
   \midrule
    \multirow{4}{*}{Concatenate} & CONCAT (Pre)  &  44.16 $\pm$ 0.03 &  65.5 $\pm$ 0.1 & 49.98 $\pm$ 0.18  &  52.85 +- 0.34  \\
     & CONCAT (e2e)  &  44.09 $\pm$ 0.08 &  66.59 $\pm$ 0.04 &  50.17 $\pm$ 0.04  & 53.23 $\pm$ 0.41  \\
     & CONCAT (e2e \& KD)    &  44.19 $\pm$ 0.05 & 66.53 $\pm$ 0.13 &  50.11 $\pm$ 0.09 & 53.66 +- 0.32 \\
    \midrule
     \multirow{4}{*}{Modulate} & FiLM (Pre) &  44.12 $\pm$ 0.09 &  65.78 $\pm$ 0.06 & 50.0 $\pm$ 0.09  & 53.16 $\pm$ 0.51  \\
     & FiLM (e2e) &  44.31 $\pm$ 0.08 &  66.33 $\pm$ 0.33 &   50.16 $\pm$ 0.05 & 53.23 $\pm$ 0.58  \\
     & FiLM (e2e \& KD)  &  44.1 $\pm$ 0.09 &  66.6 $\pm$ 0.21 &  50.31 $\pm$ 0.19 &  53.08 $\pm$ 0.3 \\
    \midrule 
    \multirow{2}{*}{HyperSVM} &
    HyperSVM (e2e)  &  43.19 $\pm$ 0.02 & 66.05 $\pm$ 0.06 & 49.78 $\pm$ 0.23 &  50.04 $\pm$ 0.31 \\
    &  HyperSVM (e2e \& KD)    &   42.55 $\pm$ 0.03 &   66.07 $\pm$ 0.05 &   49.53 $\pm$ 0.13 & 49.34 +- 0.43  \\
    \midrule
    \multirow{2}{*}{HyperSVM-FiLM} & HyperSVM-FiLM (e2e)  &   43.64 $\pm$ 0.18 & 65.54 $\pm$ 0.32 & 49.81 $\pm$ 0.27 & 50.17  $\pm$ 0.58 \\
    & HyperSVM-FiLM (e2e \& KD)    &  43.28 $\pm$ 0.48 & 65.52 $\pm$ 0.34 &  49.73 $\pm$ 0.05 &  49.73 +- 0.39 \\
    \midrule
    \midrule
     &  FiLM (e2e \& KD \& AT)  & \textbf{44.56 $\pm$ 0.13} & \textbf{66.85 $\pm$ 0.28} & \textbf{50.37 $\pm$ 0.03} & \textbf{53.68 $\pm$ 0.23 } \\
    \bottomrule
    \end{tabular}
    \label{tab:test-perf}
\end{table}

%% file: inputs/hypersvm-comparison.tex
\begin{table}[h]
    \scriptsize
    \caption{FiLM architecture generalizes to larger instances better than the HyperSVM types of architectures. We compare the performance of the best FiLM architecture from the Table~\ref{tab:test-perf} with the best HyperSVM architecture in the same table.}
    \begin{subtable}{1\textwidth}
    \centering
    \begin{tabular}{p{12mm}|rrr|| rrr|| rrr}
    \toprule
    {} & \multicolumn{3}{c}{small} & \multicolumn{3}{c}{medium} & \multicolumn{3}{c}{big} \\
    {} &   Time  &     Wins &     Nodes &       Time  &     Wins &     Nodes &  Time  &     Wins &     Nodes \\
    \midrule
    FiLM &    \textbf{24.67} &   \textbf{53}/ 60 &    \textbf{109} &   \textbf{136.42} &   \textbf{51}/ 60 &    \textbf{336} &   \textbf{531.70} &   \textbf{46}/ 57 &    \textbf{345} \\
    HyperSVM &    27.26 &    7/ 60 &    110 &   158.97 &    9/ 60 &    345 &   614.36 &   11/ 57 &    346 \\
    MLP &    27.61 &    4/ 60 &    114 &   156.30 &   11/ 60 &    347 &   595.31 &    9/ 56 &    334 \\
    \bottomrule
    \end{tabular}
    \caption{Capacitated Facility Location}
    \end{subtable}
    
    \begin{subtable}{1\textwidth}
    \centering 
    \begin{tabular}{p{20mm}|rrr|| rrr|| rrr}
    \toprule
    {} & \multicolumn{3}{c}{small} & \multicolumn{3}{c}{medium} & \multicolumn{3}{c}{big} \\
    {} &   Time  &     Wins &     Nodes &       Time  &     Wins &     Nodes &  Time  &     Wins &     Nodes \\
    \midrule
    FiLM  &     \textbf{8.73} &   \textbf{59}/ 60 &    \textbf{147} &    \textbf{63.75} &   \textbf{60}/ 60 &  \textbf{2169} &  \textbf{1843.24} &   \textbf{24}/ 26 &  \textbf{38530} \\
    HyperSVM-FiLM &     9.73 &    1/ 60 &    148 &    72.53 &    0/ 60 &   2217 &  2061.56 &    1/ 22 &  47277 \\
    MLP &     9.98 &    0/ 60 &    157 &    77.48 &    0/ 60 &   2299 &  1984.26 &    1/ 24 &  40188 \\
    \bottomrule
    \end{tabular}
    \caption{Set Cover}
    \end{subtable}
    \label{tab:hypersvm-evaluation}

\end{table}

%% file: inputs/scale-out.tex
\sisetup{detect-weight=true,detect-inline-weight=math,detect-mode=true}
\begin{table}[htb]
\caption{Performance of branching strategies on twice the size of biggest instances (2 $\times$ Big) considered in the main paper. 20 "Bigger" instances were solved using 3 seeds each resulting in a total of 60 runs. We see that FiLM models still remain competitive, and it is highly dependent on the family of problem.}
\label{tab:scale-out}
\centering
\scriptsize
\setlength{\tabcolsep}{15pt}
\aboverulesep = 0.1mm  
\belowrulesep = 0.2mm  
\begin{tabular}{
    c
    S[table-format=3.2]@{\hspace{15pt}}
    S[table-format=2.0]@{\hspace{4pt}/\hspace{1pt}}
    S[table-format=2.0]@{\hspace{5pt}}
    S[table-format=3.0]@{}
    S[table-format=3.2]@{\hspace{15pt}}
    S[table-format=2.0]@{\hspace{4pt}/\hspace{1pt}}
    S[table-format=2.0]@{\hspace{5pt}}
    S[table-format=3.0]@{}
    }
    \toprule
     &
    \multicolumn{4}{ c }{facilities} &
    \multicolumn{4}{ c }{setcover} \\

    Model &
    \multicolumn{1}{ c }{Time} &
    \multicolumn{2}{ c }{Wins} &
    \multicolumn{1}{ c }{Nodes} &
    \multicolumn{1}{ c }{Time} &
    \multicolumn{2}{ c }{Wins} &
    \multicolumn{1}{ c }{Nodes} \\
    \toprule

    \textsc{rpb}
        &   7200.14 &   0 &  0 &  n/a
        & \B 6346.17 &  \B 9 &  12 &  135132 \\

    \textsc{gnn}
        &   7111.83  &   1 & 5 &  n/a
        &  7200.25 &   0 & 0 &  n/a \\

    \textsc{FiLM} (ours)
        & \B    7052.27  & \B  4 &  4 &  n/a
        &   6508.78 & 3 &  10 &  \B 107187 \\

    \cmidrule(lr){1-5} \cmidrule(lr){6-9}
    \textsc{gnn-gpu}
        &     6625.55  &     \text{--} &  13 &  n/a 
        &     6008.14 &     \text{--} & 12 & 93909 \\
        
    \bottomrule
\end{tabular}
\vspace{-15pt}
\end{table}

%% file: inputs/training-protocol-performance.tex
\sisetup{detect-weight=true,detect-inline-weight=math,detect-mode=true}
\begin{table*}[t]
\caption{
Effect of training protocols on the performance of \textit{branching strategies}. We report geometric mean of solving times, number of times a method won (in solving time) over total finished runs, and geometric mean of number of nodes. Refer to section 5 (main) for more details. The best performing results are in \textbf{bold}. $^*$Models were regularized to prevent overfitting on small instances.}
\label{tab:protocol-evaluation}
\centering
\scriptsize
\setlength{\tabcolsep}{10pt}
\aboverulesep = 0.1mm  
\belowrulesep = 0.2mm  
\begin{tabular}{
    c
    S[table-format=3.2]@{\hspace{10pt}}
    S[table-format=2.0]@{\hspace{4pt}/\hspace{-4pt}}
    S[table-format=2.0]@{\hspace{5pt}}
    S[table-format=3.0]@{}
    S[table-format=3.2]@{\hspace{10pt}}
    S[table-format=2.0]@{\hspace{4pt}/\hspace{-4pt}}
    S[table-format=2.0]@{\hspace{5pt}}
    S[table-format=3.0]@{}
    S[table-format=3.2]@{\hspace{10pt}}
    S[table-format=2.0]@{\hspace{4pt}/\hspace{-4pt}}
    S[table-format=2.0]@{\hspace{5pt}}
    S[table-format=3.0]@{}
    }
    \toprule
     &
    \multicolumn{4}{ c }{Small} &
    \multicolumn{4}{ c }{Medium} &
    \multicolumn{4}{ c }{Big} \\

    Model &
    \multicolumn{1}{ c }{Time} &
    \multicolumn{2}{ c }{Wins} &
    \multicolumn{1}{ c }{Nodes} &
    \multicolumn{1}{ c }{Time} &
    \multicolumn{2}{ c }{Wins} &
    \multicolumn{1}{ c }{Nodes} &
    \multicolumn{1}{ c }{Time} &
    \multicolumn{2}{ c }{Wins} &
    \multicolumn{1}{ c }{Nodes} \\
    \toprule


    \text{e2e}
        &   25.05 &   22 & 60 &  114
        &  154.12 &   4 & 60  &  340
        &  550.03 &   15 & 57 &  339  \\

    \text{e2e + KD}
        &   28.00 &   2 & 60 &   111
        &  143.12 &  18 & 60 &   343
        &  \B 507.50 &  \B 26 & 57 & \B 325  \\

    \text{e2e + KD + AT}
        &  \B 24.67 &   \B 36 & 60 & \B 109
        &  \B 136.42 &  \B 38 & 60 &  \B 336
        &  531.70 &  16 & 57 &  345 \\

    \cmidrule(lr){1-5} \cmidrule(lr){6-9} \cmidrule(lr){10-13}
    \\[-7pt]
    & \multicolumn{12}{ c }{Capacitated Facility Location} \\
    \\[-3pt]

    \cmidrule(lr){1-5} \cmidrule(lr){6-9} \cmidrule(lr){10-13}

    \text{e2e}
        &  9.70   &  1 & 60 &  152 
        &  71.18  &  1 & 60 &  2186
        &  1869.57 & 4 & 25 & \B  40341  \\

    \text{e2e + KD}
        &   9.81 &    0 & 60 &   \B 146
        &  70.88 &    4 & 60 &   2173
        &  \B 1842.29 & \B 15 & 25 &  40437  \\

    \text{e2e + KD + AT}
        &  \B 8.73 &  \B 59 & 60 &   147
        &  \B 63.75 & \B 55 & 60 & \B  2169
        &  1843.24 & 8 & 26  &  40881 \\

    \cmidrule(lr){1-5} \cmidrule(lr){6-9} \cmidrule(lr){10-13}

    \\[-7pt]
    & \multicolumn{12}{ c }{Set Covering} \\
    \\[-3pt]

    \cmidrule(lr){1-5} \cmidrule(lr){6-9} \cmidrule(lr){10-13}

    \text{e2e}
        &   2.45 &    1 & 60 &  \B  72
        &  17.59 &    5 & 60 &   702
        &  225.88 &  17 & 60 &   8939  \\

    \text{e2e + KD}
        &   2.35  &    0 & 60  &  73
        &  17.59  &    2 & 60 &    720
        &  241.95 &    7 & 59 &   8846  \\

    \text{e2e + KD + AT}
        & \B  2.13 &  \B 59 & 60 &    73
        & \B 15.71 &  \B  53 & 60 & \B  686
        & \B 217.02 & \B  36 & 60 & \B  8711 \\

    \cmidrule(lr){1-5} \cmidrule(lr){6-9} \cmidrule(lr){10-13}

    \\[-7pt]
    & \multicolumn{12}{ c }{Combinatorial Auctions} \\
    \\[-3pt]

    \cmidrule(lr){1-5} \cmidrule(lr){6-9} \cmidrule(lr){10-13}

    \text{e2e}
        &   205.63  &  2 & 54  &    559
        &  1103.47 &  1 & 29 &   1137
        &  2457.94 &  1 &  4 &  \B 2268  \\

    \text{e2e + KD}
        &   333.52 &   1 & 52 &    486
        &  926.12 &    1 & 41 &    604
        &  2503.65 &  0 &  7 &   1953  \\

    \text{e2e + KD + AT}
        & \B 52.96 &  \B 54 & 55 &   \B 410
        & \B 131.45 & \B 54 & 54 &  \B  331
        & \B 1823.29 & \B 14 & 15 &   3049 \\
     
    \cmidrule(lr){1-5} \cmidrule(lr){6-9} \cmidrule(lr){10-13}
    \\[-7pt]
     & \multicolumn{12}{ c }{Maximum Independent Set$^{*}$} \\
     \bottomrule
\end{tabular}
\end{table*}